\def\year{2019}\relax
\documentclass[letterpaper]{article} 
\usepackage{aaai19}  
\usepackage{times}  
\usepackage{helvet}  
\usepackage{courier}  
\usepackage{url}  
\usepackage{graphicx}  
\usepackage{amsthm,amsmath,mathrsfs,amssymb}
\usepackage{epstopdf}
\usepackage{bbm}
\usepackage{color}
\usepackage{comment}
\usepackage{multirow}
\usepackage{bm}
\usepackage{float}
\usepackage{stfloats}
\usepackage[tight]{subfigure}
\usepackage{graphicx,color,url}
\usepackage{subfigure}

\usepackage{keyval}
\usepackage{algorithm}
\usepackage{algorithmic}
\usepackage{tabularx,booktabs}
\usepackage[colorlinks, citecolor=blue]{hyperref}
\usepackage{extarrows}
\usepackage{threeparttable}

\usepackage{float}

\newsavebox{\tempbox}

\newcommand{\RomanNumeralCaps}[1]
    {\MakeUppercase{\romannumeral #1}}

\newtheorem{theorem}{Theorem}
\newtheorem{lemma}{Lemma}
\newtheorem{definition}{Definition}

\newtheorem{proposition}{Proposition}
\newtheorem{corollary}{Corollary}

\begin{document}
%
\title{Geometric Operator Convolutional Neural Network}
\author{Yangling Ma \quad Yixin Luo \quad Zhouwang Yang \\
School of Mathematical Sciences, USTC}
\maketitle
\begin{abstract}
The Convolutional Neural Network (CNN) has been successfully applied in many fields during recent decades; however it lacks the ability to utilize prior domain knowledge when dealing with many realistic problems. We present a framework called Geometric Operator Convolutional Neural Network (GO-CNN) that uses domain knowledge, wherein the kernel of the first convolutional layer is replaced with a kernel generated by a geometric operator function. This framework integrates many conventional geometric operators, which allows it to adapt to a diverse range of problems.

Under certain conditions, we theoretically analyze the convergence and the bound of the generalization errors between GO-CNNs and common CNNs. Although the geometric operator convolution kernels have fewer trainable parameters than common convolution kernels, the experimental results indicate that GO-CNN performs more accurately than common CNN on CIFAR-10/100. Furthermore, GO-CNN reduces dependence on the amount of training examples and enhances adversarial stability. In the practical task of medically diagnosing bone fractures, GO-CNN obtains 3\% improvement in terms of the recall.
\end{abstract}


\section{Introduction}\label{sec:intro}
Convolutional Neural Networks have been successfully applied in many fields during recent decades, but the theoretical understanding of the deep neural network is still in the preliminary stages. Although Convolutional Neural Networks have strong expressive abilities, they have to two clear deficiencies. First, as complex functional mappings, Convolutional Neural Networks, like black boxes, cannot take full advantage of domain knowledge and prior information. Second, when little data is available for a certain task, Convolutional Neural Networks' generalization ability weakens. This is due to overfitting, which may occur due to the large number of parameters and the large model size. Stemming from these two defects, a great deal of research has been done to modify CNNs \cite{dai2017deformable} \cite{wang2018non} \cite{sarwar2017gabor}.

Before CNNs were applied, traditional geometric operators had developed quite well. Each geometric operator represents the precipitation of domain knowledge and prior information. For example, the Sobel operator \cite{workssobel} is a discrete difference operator, which can extract image edge information for edge detection. The Schmid operator \cite{schmid2001constructing} is an isotropic circular operator, which extracts texture information from images for face recognition. The Histogram of Oriented Gradients (HOG) \cite{dalal2005histograms} is a statistic operator of gradient direction, which extracts edge direction distributions from images for pedestrian detection and other uses.

 Many computer vision tasks require domain knowledge and prior information. Geometric operators can make use of domain knowledge and prior information, but cannot automatically change parameter values by learning from data. Convolutional Neural Networks have strong data expression abilities and learning abilities, but they struggle to make use of domain knowledge. For better data learning, we have combined the two. It is natural to directly use geometric operators for pre-processing, and then classify the data through a Convolutional Neural Network \cite{yao2016gabor}. However, this method uses human experience to select geometric operator parameter values, and then carries out the Convolutional Neural Network learning separately. This method is a kind of two-stage technique, and without reducing parameter redundancy in a Convolutional Neural Network, it is difficult to achieve global optimization. The method proposed in this paper directly constructs geometric operator convolution and then integrates geometric operator convolution into a Convolutional Neural Network to form a new framework - the Geometric Operator Convolutional Neural Network. This method achieves global optimizations and utilizes the properties of geometric operators.

~\\
In summary, the contributions of this work are as follows:

\begin{itemize}

\item This framework can integrates many conventional geometric operators, which reveals its broad customization capabilities when handling diverse problems.

\item In theory, the same approximation accuracy and generalization error bounds are achieved when geometric operators meet certain conditions.

\item The Geometric Operator Convolutional Neural Network not only reduces the redundancy of the parameters, but also reduces the dependence on the amount of the training samples.

\item The Geometric Operator Convolutional Neural Network enhances adversarial stability.

\end{itemize}

We organize the remaining chapters in the following sequence. We first briefly introduce related work in Sec. \ref{sec:relate}. In Sec. \ref{sec:method}, we describe our framework for the Geometric Operator Convolutional Neural Network, and in Sec. \ref{sec:them}, we introduce theoretical analyses. Experiments and conclusion are presented in Sec. \ref{sec:expe}, Sec. \ref{sec:con}, respectively.


\section{Related Work}\label{sec:relate}
In recent years, Convolutional Neural Networks have been widely used in various classification and recognition applications \cite{krizhevsky2012imagenet} \cite{hu2014convolutional}. Convolutional Neural Networks have achieved advanced success in various problems. All CNNs adopt an end-to-end approach to learning; however, each unique task is associated with its own distinctive domain knowledge and prior information. Thus, to improve classification accuracy, researchers use priori information that is tailored to each specific task and each specific Convolutional Neural Network. One way to do this is to use the traditional image processing algorithm as a preprocessing step. Another way is to use the traditional image processing algorithm to initialize convolution kernels.

Classification accuracy is a primary concern for researchers in the machine-learning community. Different pre-processing models, such as filters or feature detectors, have been employed to improve the accuracy of CNNs. One example of this is the Gabor filter with CNN \cite{daugman1988complete}. The Gabor filter is a feature extractor based on human vision. Besides the Gabor filter, some people also use Fisher vectors \cite{cimpoi2014deep}, sparse filter Banks \cite{pfister2015learning}, and the HOG algorithm \cite{lu2018fast} combined with a CNN to improve accuracy.  Based on the human visual system, these filters are found to be remarkably well-suited for texture representation and discrimination. In the works by Bogdan et al. \cite{kwolek2005face} and Mounika et al. \cite{mounika2012neural}, the Gabor filter is used to extract features from the input image in a pre-processing step. However, these methods require a kind of two-stage procedure that may not reach the optimal global solution.

In addition, some scholars use traditional image processing algorithms to initialize convolutional kernels, such as building a Feature Pyramid Network with an image pyramid for multi-scale feature extraction \cite{lin2017feature}. Geometric operators are widely used in traditional image processing algorithms. Many researchers use the Gabor filter to fix the first convolution layer, while other layers, which are common convolution layers, can be trained to improve their accuracy \cite{yao2016gabor} \cite{sarwar2017gabor}. Vijay et al. \cite{johngabor} simultaneously adopted the weight of the first layer convolution with the Gershgorin circle theorem and the Gabor filter constraint to improve the classification accuracy when Convolutional Neural Networks propagated backward. In \cite{calderon2003handwritten} \cite{chang2014robust}, the authors have attempted to get rid of the pre-processing overhead by introducing Gabor filters in the first convolutional layer of a CNN. In addition, some researchers use filters to initialize multiple convolutional kernels. Shangzhen et al. \cite{lu2018fast} only used the Gabor function to create kernels in four directions to initialize the convolutional kernels from a Convolutional Neural Network. These methods change the initialization weight and use domain knowledge, but they do not reduce the redundancy of model parameters, and they do not enhance the transformation ability of the model.

In summary, the above two ways only use domain knowledge and prior information to improve Convolutional Neural Networks and classification accuracy. In this paper, a new network, the Geometric Operator Convolutional Neural Network, is proposed. This method integrates geometric operators, namely the filters, into a convolutional neural network. This network can not only make use of domain knowledge and prior information, but also reduce the redundancy of network parameters and enhance the ability of model transformation.


\section{The framework of the Geometric Operator Convolutional Neural Network}\label{sec:method}

Traditional geometric operators have many properties. By convolving geometric operators and integrating them into a deep neural network to form a Geometric Operator Convolutional Neural Network, we not only retain the characteristics of geometric operators, but also give play to the powerful feature expression ability of deep neural networks. This framework renders image classification tasks more effective. The method's construction is described in detail in the following section.


\subsection{Geometric operators}\label{sec:method:diveraity}

Before the development of deep Convolutional Neural Networks, traditional image feature extraction methods were based on traditional image processing algorithms, primarily geometric operators. At present, a large number of geometric operators have been applied, such as the Scale Invariant Feature Transform (SIFT) \cite{lowe1999object}, the Roberts operator \cite{rosenfeld1981max}, the Laplace operator \cite{van1989nonlinear}, the Gabor operator \cite{han2007rotation}, and so on. Each operator has different characteristics. Therefore, different geometric operators are used in different application scenarios, according to the characteristics of each unique problem. For example, SIFT looks for feature points in different scale spaces for pattern recognition and image matching. The Roberts operator uses local differences to find edges for edge detection, and the Laplace operator uses isotropic differentials to retain details for image enhancement. 

Geometric operators represent the precipitation of domain knowledge and prior knowledge. The Geometric Operator Convolutional Neural Network is proposed in this paper, which uses the characteristics of geometric operators. The first step in this framework is to convolve geometric operators. In this paper, the Gabor operator and the Schmid operator are mainly used as examples to illustrate how to carry out convolutions and integrate these convolutions into Convolutional Neural Networks. Other geometric operators in subsequent studies employ similar concepts.


\subsection{Convolution of geometric operators}\label{sec:method:para}

\subsubsection{Gabor operator}\label{sec:method:para:gabor}
\begin{figure}[!htb]
\centering
\subfigure[Convolution kernels from the first layer of ResNet50]{
\label{fig:2}
\includegraphics[width=0.4\textwidth]{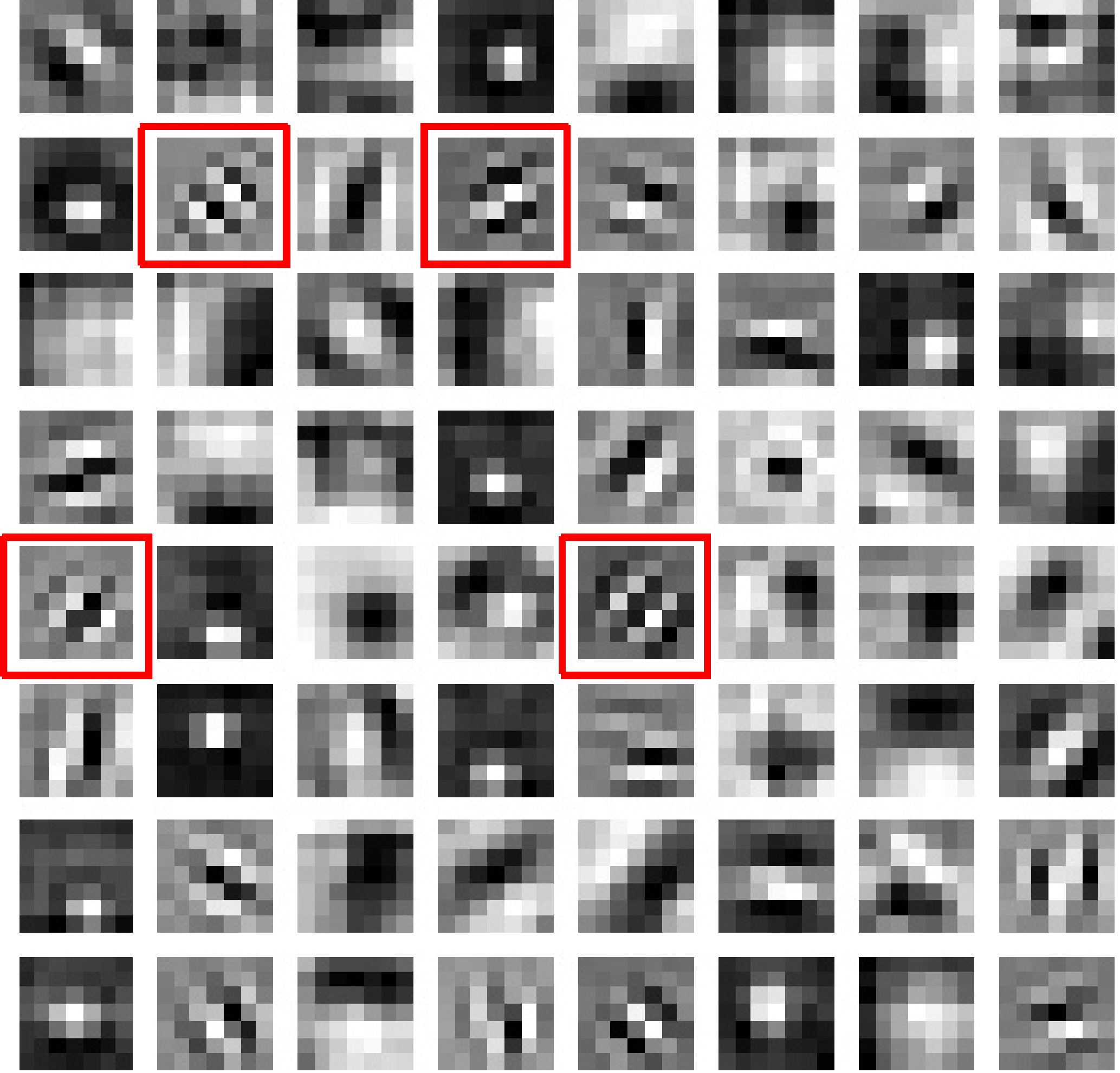}}
\quad
\subfigure[Gabor kernels]{
\label{fig:3}
\includegraphics[width=0.4\textwidth]{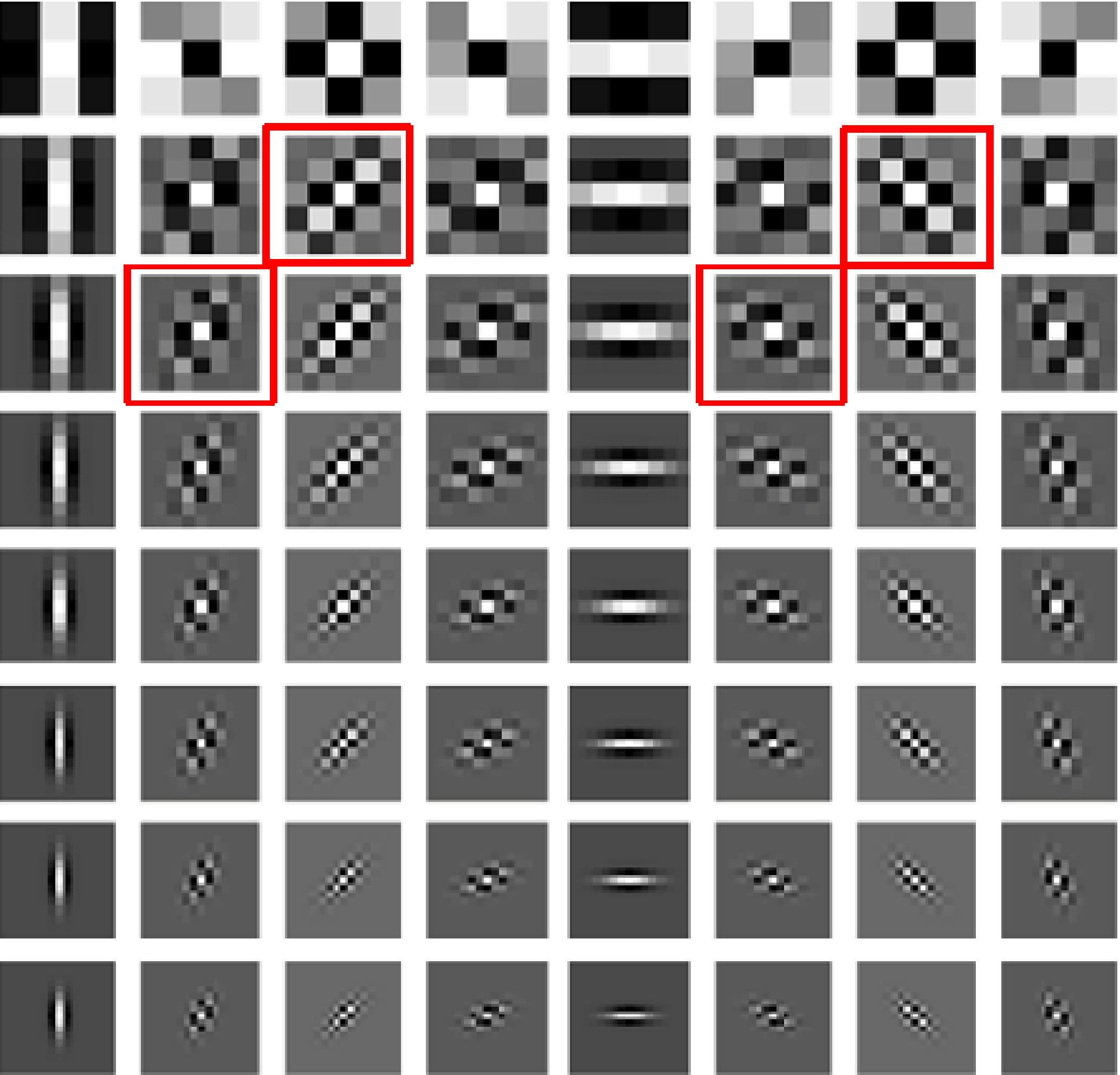}}
\caption{The similarities between the CNN's first convolutional kernels and Gabor kernels.}
\label{fig:sim}
\end{figure}

In order to study the frequency characteristics of local range signals, Dennis Gabor \cite{gabor1946theory} proposed the famous ``Window" Fourier transform (also called the short-time Fourier transform, STFT) in the paper ``Theory of communication" in 1946. This is now known as the Gabor operator; when combined with images, it is referred to as the Gabor filter. Until now, the Gabor filter has undergone many developments, and its primary characteristics are listed below. First, the Gabor filter has the advantages of both spatial and frequency signal processing. As shown in Eqn. \ref{eqn:1}, the Gabor operator is essentially a Fourier transform with a gaussian window. For an image, the window function determines its locality in the spatial domain, so the spatial domain information from different positions can be obtained by moving the center of the window. In addition, since the gaussian function remains the same after the Fourier transform, the Gabor filter can extract local information in the frequency domain. Second, the Gabor filter's response to biological visual cells may be an optimal feature extraction method. In 1985, Daugman \cite{daugman1985uncertainty} extended the Gabor function to a 2-dimensional form and constructed a 2D Gabor filter on this basis. It was surprising to find that the 2D Gabor filter was also able to obtain both the minimum uncertainty of time and frequency domain at the same time, while maintaining consistency with the mammalian model of retinal nerve cell reception. Third, the Gabor kernels are similar to the convolution kernels from the first convolutional layer in the CNN. An illustration of this similarity is shown in Fig. \ref{fig:sim}. From the visualization of the first convolutional layer in AlexNet, which was proposed by Alex et al. \cite{krizhevsky2012imagenet}. Some convolution kernels present geometric properties, as in the kernel function from the Gabor filter. From this feature, it can also be explained that there are parameter redundancies in the Convolutional Neural Network, and the Gabor operator can be convoluted and integrated into CNN. Lastly, the Gabor filter can extract directional correlation texture features from an image. As shown in Fig. \ref{fig:4}, there are 40 Gabor kernels from five scales and eight directions convolving with an image. Texture feature maps in different directions can be obtained from the original image.  

\begin{equation}
\begin{split}
g_{\theta, \phi, \gamma, \sigma, \lambda}(x, y) =&\ exp\left(-\frac{{x'}^2+{y'}^2}{2\sigma^2}\right)cos\left(\frac{2\pi x'}{\lambda} + \phi\right) \\
x' =&\ xcos\theta + ysin\theta \\
y' =&\ -xsin\theta + ycos\theta
\end{split}
\label{eqn:1}
\tag{1. 0}
\end{equation}


\begin{figure}[!htb]
\centering
\includegraphics[width=0.5\textwidth]{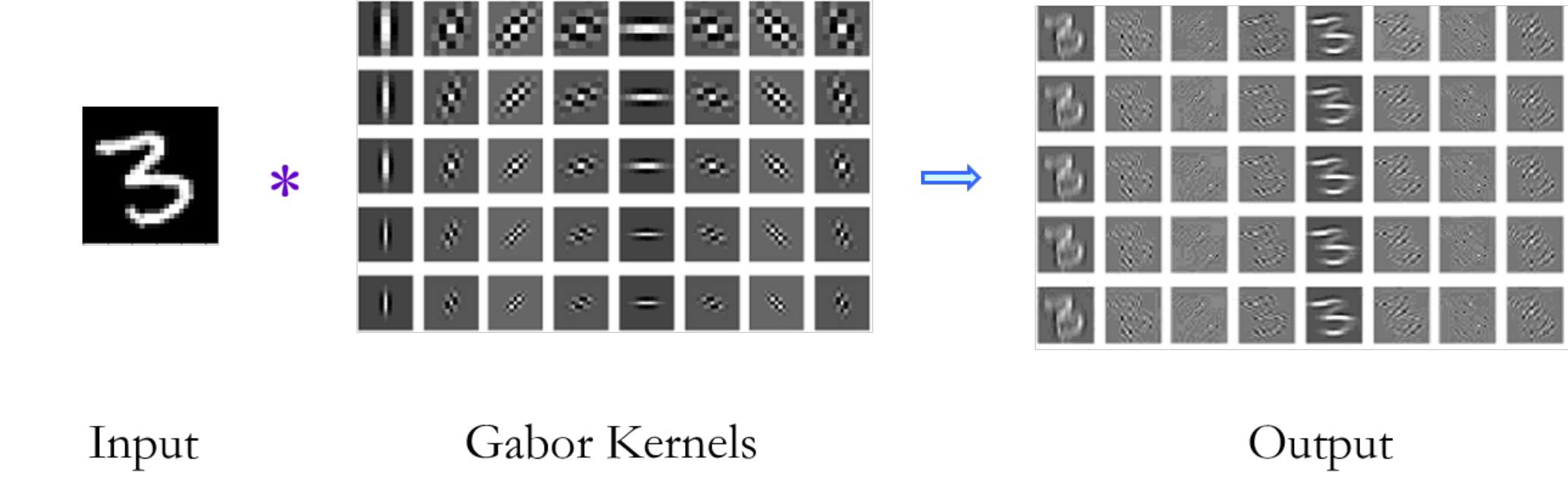}
\caption{The results of the Gabor operator on an image}\label{fig:4}
\label{fig:4}
\end{figure}

Since the Gabor operator combines with the CNN in the image, better feature expressions can be obtained. There are two main binding methods. First, the image is preprocessed by the Gabor operator, and then its features are extracted by the CNN. Next, the gabor operator is convoluted to form a convolution layer, and then we integrate this convolution into the common Convolutional Neural Network. The second approach is used in this article. As shown in Eqn. \ref{eqn:1}, the Gabor kernel function has 5 parameters, which are obtained by learning and then regenerated into an $m \times m$ kernel. We replace the common convolution kernels with these Gabor kernels to form a convolutional layer. However, for the common convolutional layer, an $m \times m$ convolution kernel is generated by an identity mapping, which requires $m^{2}$ parameters. So, our method reduces the number of trainable parameters in the convolutional layer.

\subsubsection{Schmid operator}\label{sec:method:para:schmid}

In 2001, Schmid et al. \cite{schmid2001constructing} proposed a Gabor-like image filter, namely the Schmid operator. As shown in Eqn. \ref{eqn:2}, its composition is similar to the kernel function of the Gabor operator, so it retains the properties of the Gabor operator. In addition, as shown in Fig. \ref{fig:5}, when the original image and a version of that image that has been rotated 90 degrees are both convolved with the same Schmid kernel, the resulting characteristic graph exhibits only 90 degrees of rotation; in other words, the Schmid operator has rotation invariance. The Schmid operator is then convoluted, and we integrate this convolution into common Convolutional Neural Network. This network improves the model's adversarial stability to rotation and improve the image feature extraction effect. Similar to the convolution of the Gabor operator, as shown in Eqn. \ref{eqn:2}, the Schmid kernel function has two parameters, which are obtained by learning and then generated by the Schmid kernel. Finally, we replace common convolution kernels with Schmid kernels to form a convolutional layer.

\begin{figure}[!htb]
\centering
\includegraphics[width=0.5\textwidth]{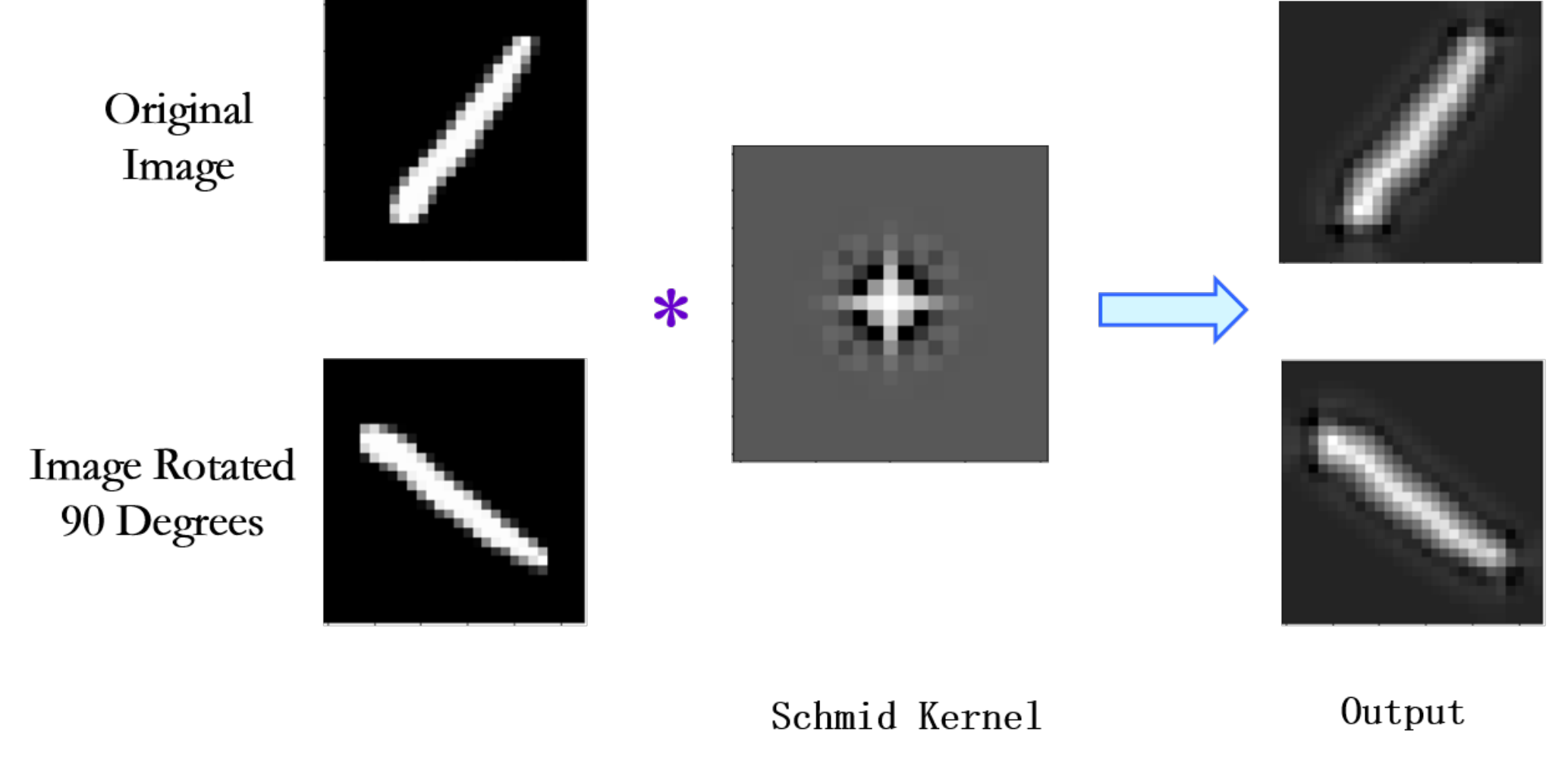}
\caption{The results of the Schmid operator on an image}\label{fig:5}
\label{fig:5}
\end{figure}

\begin{equation}
\begin{split}
F_{\sigma, \tau}(r) =&\ exp\left(-\frac{r^2}{2\sigma^2}\right)cos\left(\frac{2\pi \tau r}{\sigma}\right) \\
r =&\ \sqrt{x^2 + y^2}
\end{split}
\label{eqn:2}
\tag{2. 0}
\end{equation}

In this paper, only two geometric operator convolutions are explained. Similarly, for other geometric operators, operator kernels are generated by operator kernel functions, which replace common convolution kernels to form a convolutional layer. Due to the diversity of geometric operators, different geometric operators can be replaced with geometric operator convolutions, so the geometric operator convolution is customizable. There is a kind of geometric operator to form any kind of geometric operator convolution. Consequently, a question that must be addressed is how we combine multiple geometric operators with common Convolutional Neural Networks to form the Geometric Operator Convolutional Neural Network.


\subsection{Geometric Operator Convolutional Neural Network}\label{sec:method:frame}
\begin{figure}[!htb]
\centering
\includegraphics[width=0.5\textwidth]{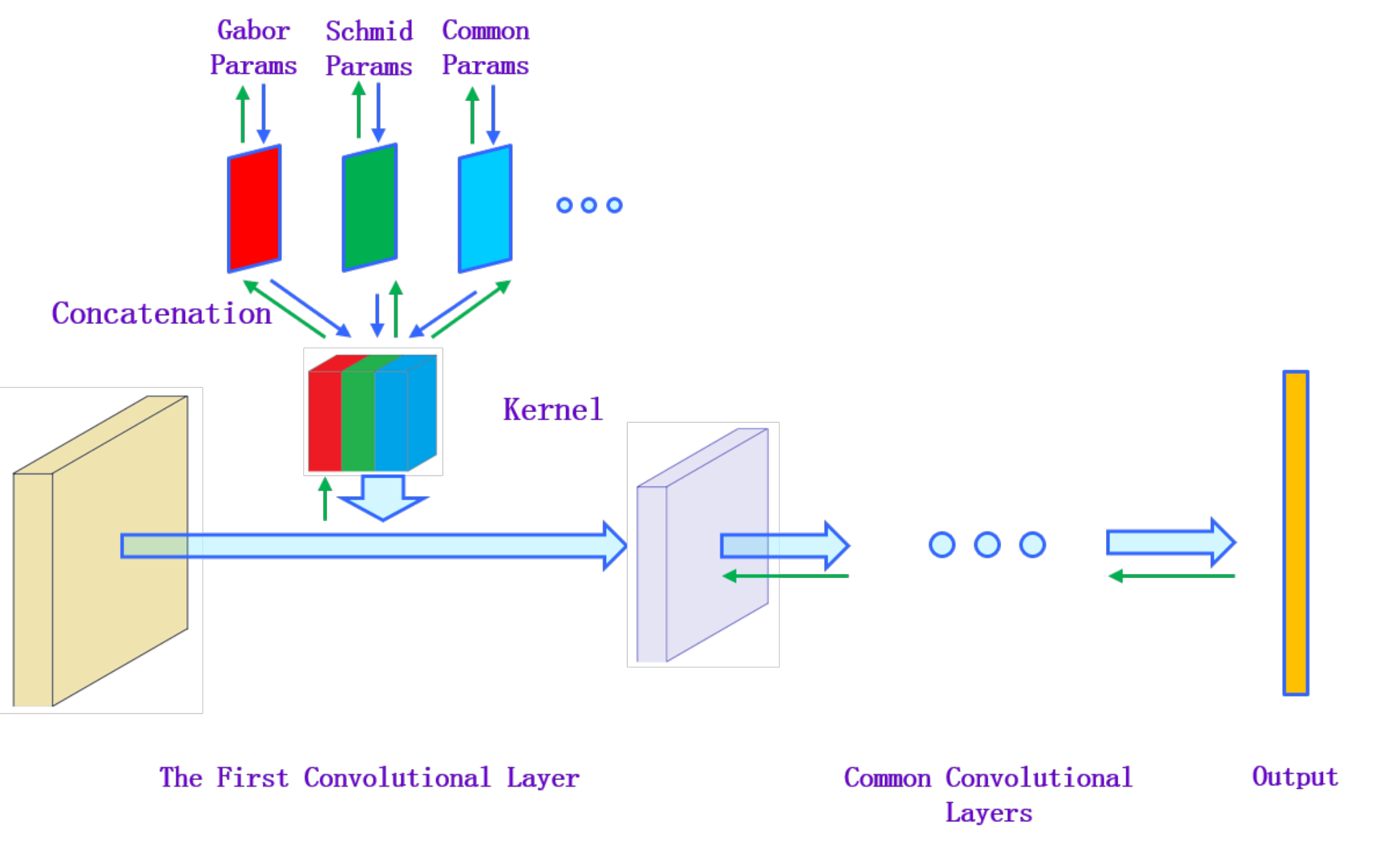}
\caption{The architecture of our framework}
\label{fig:6}
\end{figure}

Only a visualization of the first layer of convolution kernels maintains some geometric characteristics, so the Geometric Operator Convolutional Neural Network proposed in this paper only replaces kernels from the first convolutional layer with geometric operator kernels. The framework of the Geometric Operator Convolutional Neural Network is introduced in Fig. \ref{fig:6}. First, kernels from the first convolutional layer are calculated by the parameters of various geometric operators. Then, we concatenate all the calculated convolutional kernels in the last dimension to obtain a complete convolutional kernel. This convolution kernel is used as the weight of the first convolution layer in the Geometric Operator Convolutional Neural Network, and then the common convolution layer and output layer are connected. In this way, we have defined the forward propagation of the whole Geometric Operator Convolutional Neural Network. So, in reverse propagation, the gradient of loss is transferred to the convolution kernel; this process is different from the usual convolution. Here, the convolution kernel generated by the geometric operator needs to further use the chain derivative rule (i. e., Eqn. \ref{eqn:3}, where $L$ is the loss function, $w$ is each convolution kernel, and $p_i$ is the parameter to generate each convolution kernel) to transfer the gradient to the parameters of each convolution kernel. Then, trainable parameters are updated by gradient descent algorithms, and the whole Geometric Operator Convolutional Neural Network is complete.

\begin{equation}
\begin{split}
\frac{\partial L}{\partial p_i} =& \frac{\partial L}{\partial w}\frac{\partial w}{\partial p_i}
\end{split}
\label{eqn:3}
\tag{3. 0}
\end{equation}


\section{Theoretical analyses}\label{sec:them}
The whole framework of the Geometric Operator Convolutional Neural Network has been introduced above. Next, we describe how to theoretically analyze the Geometric Operator Convolutional Neural Network. It is theoretically proved that although the number of trainable parameters in the Geometric Operator Convolutional Neural Network decreases, the effectiveness for computer vision tasks does not decrease.

\subsection{\textbf{Definition of data and loss function}}
\begin{itemize}
\item We denote the input by $\mathcal{S}=\{I_i\}_{i=1}^N$, the corresponding label is $\{y_i|y_i=0\ or\ 1\}_{i=1}^N$.
\item The loss function is \textit{Mean Square Error}.
\item The output of the neural network is $\tilde{y}_i$ for each input $I_i$, and the empirical loss function is defined as follows:

\begin{equation}
\hat{\mathbb{E}}_S[h]=\frac{1}{N}\sum_{i=1}^{N}(\tilde{y}_i - y_i)^2
\tag{4.0}
\end{equation}
\end{itemize}

\begin{lemma}\cite{cybenko1989approximation}\label{sec:them:lemma1}
 Define a functional class $\Pi \subset \{f\ |\ f: \mathbb{R}^d \mapsto [0, 1]\}$, where each $f \in \Pi$ can be approximated with error at most $\epsilon$ by a one hidden-layer neural network $N$, that is:
 \begin{equation}
 |f(x)-N(x)| \leq \epsilon, \quad \forall x
 \label{eqn:lemma1}
 \tag{5.0}
 \end{equation}
\end{lemma}

\begin{lemma}\cite{bartlett2002rademacher}\label{sec:them:lemma2}
 let $\mathcal{F}$ and $\mathcal{G}$ be two hypothesis classes and let $a \in \mathbb{R}$ be a constant, we have:

\begin{equation}
\begin{split}
\mathcal{F} \subseteq \mathcal{G} \Rightarrow \hat{\mathfrak{R}}_{S}^a(\mathcal{F}) \leq \hat{\mathfrak{R}}_{S}^a(\mathcal{G}) \\
\hat{\mathfrak{R}}_{S}^a(\mathcal{F} + \mathcal{G}) \leq \hat{\mathfrak{R}}_{S}^a(\mathcal{F}) + \hat{\mathfrak{R}}_{S}^a(\mathcal{G})
\end{split}
\label{eqn:lemma2}
\tag{6.0}
\end{equation}
\end{lemma}

\begin{lemma}\cite{mohri2012foundations}\label{sec:them:lemma3}
Let $z$ be a random variable of support $\mathcal{Z}$ and distribution $\mathcal{D}$. Let $S = \{z_1, z_2, \cdots, z_N\}$ be a data set of $N$ i.i.d. samples drawn from $\mathcal{D}$. Let $\mathcal{F}$ be a hypothesis class satisfying $\mathcal{F} \subseteq \{f\ |\ f:\mathcal{Z}\rightarrow[a, a+1]\}$. Fix $\delta \in (0, 1)$. With probability at least $1-\delta$ over the choice of $S$, the following holds for all $h \in \mathcal{F}$:

\begin{equation}
\mathbb{E}_{\mathcal{D}}[h] \leq \hat{\mathbb{E}}_S[h] + 2\hat{\mathfrak{R}}_{S}^a(\mathcal{F}) + \sqrt{\frac{log(1/\delta)}{2N}}
\label{eqn:lemma3}
\tag{7.0}
\end{equation}
\end{lemma}

\begin{definition}[\textbf{Parametric Convolutional Kernel Space}]\label{sec:them:def:param}
Let $f$ be a function that maps vector from $\mathbb{R}^n$ to matrix in $\mathbb{R}^{m \times m}, \quad n, m \in \mathbb{N}^+$, and we call this function as \textit{convolution kernel generator function}. Then we define \textit{Parametric Convolutional Kernel Space} $\mathcal{K}_f$ as:

\begin{equation}
\begin{split}
\mathcal{K}_f = \lbrace [f(p^1), f(p^2), \cdots, &f(p^{od})], p^i \in D \subset \mathbb{R}^n, \\
&i=1, 2, \cdots, od \rbrace. \\
\end{split}
\tag{8.0}
\label{eqn:def:param}
\end{equation}

We call $n$ the parameter number, $m$ the kernel size, $od$ (short for output dimension) the output dimension. Since a convolutional kernel in a parametric convolutional kernel spaces is generated by function $f$, we call $f$ as the \textit{generator function}, and $f_{i, j}(p) = f(p)[i, j]$ as the \textit{pixel generator function}.

\end{definition}

Once we have defined the parametric convolutional kernel space, we can use $n$ parameters to generate $m \times m$ sized convolution kernel through the generator function $f$, which means that the parameters in a convolutional layer can be reduced if $n \ll m \times m$. However, the reduction in parameters often causes loss of performance as the hypothesis space becomes smaller. Therefore, how the reduction of parameters affects the performance is the key point. We want to study the simplest situation, that is to say, we want to replace the ordinary kernel in the first convolutional layer with the parameter kernel generated from a parametric convolutional kernel space.

\begin{definition}[\textbf{Geometric Operator CNN}]\label{sec:them:gocnn}
Assume that $\mathcal{K}_f$ is a parametric convolutional kernel space. If the kernel in the first convolutional layer of a convolutional neural network is generated from $\mathcal{K}_f$, we call this network Geometric Operator CNN. We denote the set of Geometric Operator CNN by $\mathcal{G}_f$.
\end{definition}

\textit{Geometric Operator CNN} is almost exactly the same as common CNN, except for the kernel in the first convolutional layer. We treat the first convolutional layer as a function from images to outputs, which then act as input of the following layer. If this function is not an injective function, meaning that different inputs can be mapped to identical outputs, then the network takes these identical outputs as the input of the following layers, meaning that the final outputs are still the same. However, the image inputs of the first convolutional layer are different, and corresponding labels can also be different. Thus, when the final outputs are the same, errors must occur.

Therefore, we need to choose kernel carefully to make the function be an injective function. Since the convolution operator is a linear operator, we have the following proposition.

\begin{proposition}\label{sec:them:prop1}
If the kernel of a convolutional layer, denoted by $w$, satisfies the following:

\begin{equation}
I * w = 0 \Leftrightarrow I = 0,\quad \forall I
\tag{9.0}
\label{eqn:injective}
\end{equation}

where $I$ is the layer input and $*$ is the convolution operation, then this convolutional layer is an injective function.

\end{proposition}

We find a necessary and sufficient condition for a convolutional layer to be an injective function. But which kernel satisfies this condition? In the proposition below, we show that $3 \times 3$ kernel generated by \textit{Gabor filter} function satisfies this condition.

\begin{proposition}\label{sec:them:prop2}
Let $f$ be the Gabor filter function, that is $f_{x, y}(\theta, \sigma, \gamma, \lambda, \psi) = exp(-\frac{{x'}^2 + \gamma^2 {y'}^2}{2 \sigma^2})cos(2 \pi \frac{x'}{\lambda} + \psi)$, where $x' = xcos\theta + ysin\theta, y' = -xsin\theta + ycos\theta$. Let $\mathcal{K}_f$ be the corresponding parametric convolutional kernel space with kernel size $m$ equal to 3 and sufficient output dimension $od$. Then, there exists kernel in $\mathcal{K}_f$ satisfies the condition (\ref{eqn:injective}).
\end{proposition}

As the kernel generated from $\mathcal{K}_f$ could not meet the (\ref{eqn:injective}), we have the following definition:

\begin{definition}[\textbf{Well-Defined Geometric Operator CNN}]\label{sec:them:def:wgocnn}
Let $G \in \mathcal{G}_f$, if there is a kernel generated by $K_f$ that satisfies (\ref{eqn:injective}), we call $G$ a well-defined Geometric Operator CNN. We denote the set of all well-defined Geometric Operator CNNs as $\mathcal{G}_f^*$.
\end{definition}

\begin{corollary}\label{sec:them:coro1}
If the \textit{generator function} $f$ is Gabor filter function, the Geometric Operator CNN is well-defined.
\end{corollary}

Now, let us consider a Convolutional Neural Network with one convolutional layer and two fully-connected layers, and we will study the convergency of common CNN and Geometric Operator CNN. For the common CNN, denoted by $F$, we define the convolution kernel as $k_F$. The weights of the rest of fully-connected layers are $\{a_{F, 1}, a_{F, 2}\}$, and the biases of three layers are $\{b_{F, 0}, b_{F, 1}, b_{F, 2}\}$. Let $\sigma$ stand for \textit{sigmoid activation function}, then the convolutional layer $C_F$ and the fully-connected layer $FC_F$ can be defined as follows:

\begin{equation}
\begin{split}
C_F(x) =&\ x * k_F + b_{F, 0} \\
FC_{F, k}(x) =&\  a_{F, k}x + b_{F, k}, \quad k=1, 2 \\
\end{split}
\tag{10.1}
\label{def:convfc}
\end{equation}

Then, the last two fully-connected layers can be defined as:

\begin{equation}
D_F(x) =\ FC_{F,2} \circ \sigma \circ FC_{F, 1}(x) \\
\tag{10.2}
\label{def:doublefc}
\end{equation}

Therefore, the output before activation, denoted by $F(x)$, and after activation, denoted by $\tilde{F}(x)$, are defined as:

\begin{equation}
\begin{split}
F(x) =&\ D_F \circ \sigma \circ C_F(x) \\
\tilde{F}(x) =& \ \sigma \circ F(x) \\
\end{split}
\tag{10.3}
\label{def:ordconv}
\end{equation}

We denote the set of common CNN as $\mathcal{F}$, that is, $\mathcal{F} = \{F\}$, and the output before activation and after activation of input $I_i$ as $F_i, \tilde{F}_i$.

For a Geometric Operator CNN $G$, we similarly define the convolutional kernel to be $k_g$, and the weights and biases are $\{a_{G, 1}, a_{G, 2}, b_{G, 0}, b_{G, 1}, b_{G, 2}\}$. Then, we have the following shorthand when the input is $x$:

\begin{equation}
\begin{split}
C_G(x) =&\ x * k_g + b_{G, 0} \\
FC_{G, k} =&\ a_{G, k}x + b_{G, k}, \quad k=1, 2 \\
D_G(x) =&\ FC_{G, 2} \circ \sigma \circ FC_{G, 1}(x) \\
G(x) =&\ D_G \circ \sigma \circ C_G(x) \\
\tilde{G}(x) =& \ \sigma \circ G(x) \\
\end{split}
\tag{10.4}
\label{def:geoconv}
\end{equation}

We denote the output before activation and after activation of input $I_i$ as $G_i, \tilde{G}_i$ as well.

We maintain the same neuron number for each corresponding layer in common CNN and Geometric Operator CNN, that is to say, $dim(b_{F, k}) = dim(b_{G, k}), k=0, 1, 2$, since the approximation ability is different when the neuron number is different. We define the width of each layer as $d_k = dim(b_{F, k}) = dim(b_{G, k}), k=0, 1, 2.$

Then, the empirical loss function for common CNN and Geometric Operator CNN is:

\begin{equation}
\begin{split}
\hat{\mathbb{E}}_{\mathcal{S}}[F]=\frac{1}{N}\sum_{i=1}^{N}(\tilde{F}_i-y_i)^2 \\
\hat{\mathbb{E}}_{\mathcal{S}}[G]=\frac{1}{N}\sum_{i=1}^{N}(\tilde{G}_i-y_i)^2 \\
\end{split}
\tag{10.5}
\label{def:emploss}
\end{equation}

\noindent We have the following theorem on the difference of these two loss functions.

\begin{theorem}\label{sec:them:thm1}
For any $F \in \mathcal{F}$, where $\mathcal{F}$ is the set of common CNN, if the first fully-connected layer is wide enough, the empirical loss of a well-defined Geometric Operator CNN can be that of common CNN controls. That is, for an arbitary $\epsilon > 0$, there exists $d^* \in \mathbb{N}^+$ and $G \in \mathcal{G}_f^*$, such that when $d_1 \ge d^*$, the following inequality holds:
\begin{equation}
|\hat{\mathbb{E}}_{\mathcal{S}}[G] - \hat{\mathbb{E}}_{\mathcal{S}}[F]| \leq \epsilon
\tag{11.0}
\label{eqn:thm1}
\end{equation}
\end{theorem}

\begin{theorem}\label{sec:them:thm2}
For any $F \in \mathcal{F}$, where $\mathcal{F}$ is the set of common CNN, if the first fully-connected layer is wide enough, the generalization error of a well-defined Geometric Operator CNN can be that of common Convolutional Neural Network controlled. That is, for an arbitary $\epsilon > 0$, there exists $d^* \in \mathbb{N}^+$ and $G \in \mathcal{G}_f^*$, such that when $d_1 \ge d^*$, the following inequality holds:

\begin{equation}
\hat{\mathbb{E}}_{\mathcal{D}}[G] \leq \ \hat{\mathbb{E}}_{\mathcal{S}}[F] + 2\hat{\mathfrak{R}}_{S}^a(\mathcal{F})+\sqrt{\frac{log(1/\delta)}{2N}} + \epsilon
\tag{12.0}
\label{eqn:thm2}
\end{equation}
\end{theorem}

In Theorem. \ref{sec:them:thm2}, we know that well defined Geometric Operator CNNs have almost the same generalization error as common CNNs. Therefore, we need to find which Geometric Operator CNNs are well defined.

~\\

As Geometric Operator CNN with Gabor filter function as the \textit{generator function} is well defined, we have the following corollary.

\begin{corollary}\label{sec:them:coro2}
Let $f$ be Gabor filter function, for any $F \in \mathcal{F}$, if the first fully-connected layer is wide enough, the generalization error Geometric Operator CNN $G$, which applies $f$ as the \textit{generator function} can be that of $F$ controlled. That is, for an arbitary $\epsilon > 0$, there exists $d^* \in \mathbb{N}^+$ and $G \in \mathcal{G}_f^*$, such that when $d_1 \ge d^*$, the following inequality holds:

\begin{equation}
\hat{\mathbb{E}}_{\mathcal{D}}[G] \leq \ \hat{\mathbb{E}}_{\mathcal{S}}[F] + 2\hat{\mathfrak{R}}_{S}^a(\mathcal{F})+\sqrt{\frac{log(1/\delta)}{2N}} + \epsilon
\tag{13.0}
\label{eqn:coro2}
\end{equation}
\end{corollary}

More generally, if there are many \textit{generator functions} in the first convolutional layer of a Geometric Operator CNN, when the number of kernels generated by Gabor fiter function is sufficient enough, this Geometric Operator CNN is also well defined. Therefore, we have the following corollary.

\begin{corollary}\label{sec:them:coro3}
Let $\{f_1, f_2, \cdots, f_T\}$ be the set of \textit{generator functions}. Suppose that there are $od$ convolution kernels $\{k_1, k_2, \cdots, k_{od}\}$ in the first convolutional layer of a Geometric Operator CNN, denoted by $G$, and each $k_j$ is generated by function $f_{t_j}$, where $1 \le j \le od, 1 \le t_j \le T$. If there exists $t^* \in \{1, 2, \cdots, T\}$ such that $f_{t^*}$ is Gabor filter function, and the number of kernels generated by $f_{t^*}$, denoted by $n_{t^*}$, is sufficient big enough, then $G$ is well defined, so that (\ref{eqn:thm2}) holds.
\end{corollary}


\section{Experiments}\label{sec:expe}
In the previous chapter, we give theoretical assurance for the Geometric Operator Convolutional Neural Network. The following section includes an explanation of the experiments conducted on the geometric Operator Convolutional Neural Network. All experiments are performed on a single machine with CPU Intel Core i7-7700 CPU @ 3.60GHz × 8, GPU TITAN X (Pascal), and RAM 32G.

\subsection{Approximation accuracy, generalization error, and feature visualization}\label{sec:expe:appro}
\textbf{Approximation accuracy and generalization error} Theoretical analyses ensures that the Geometric Operator Convolutional Neural Network has the same approximation accuracy and the same upper bound for generalization error as the common Convolutional Neural Network. We verify this using two kinds of experiments on CIFAR-10/100. The generalization error refers to the performance of the model on the test set, and the approximation accuracy refers to the performance of the model on the training set.

\begin{figure}[!htb]
\centering
\includegraphics[width=0.5\textwidth]{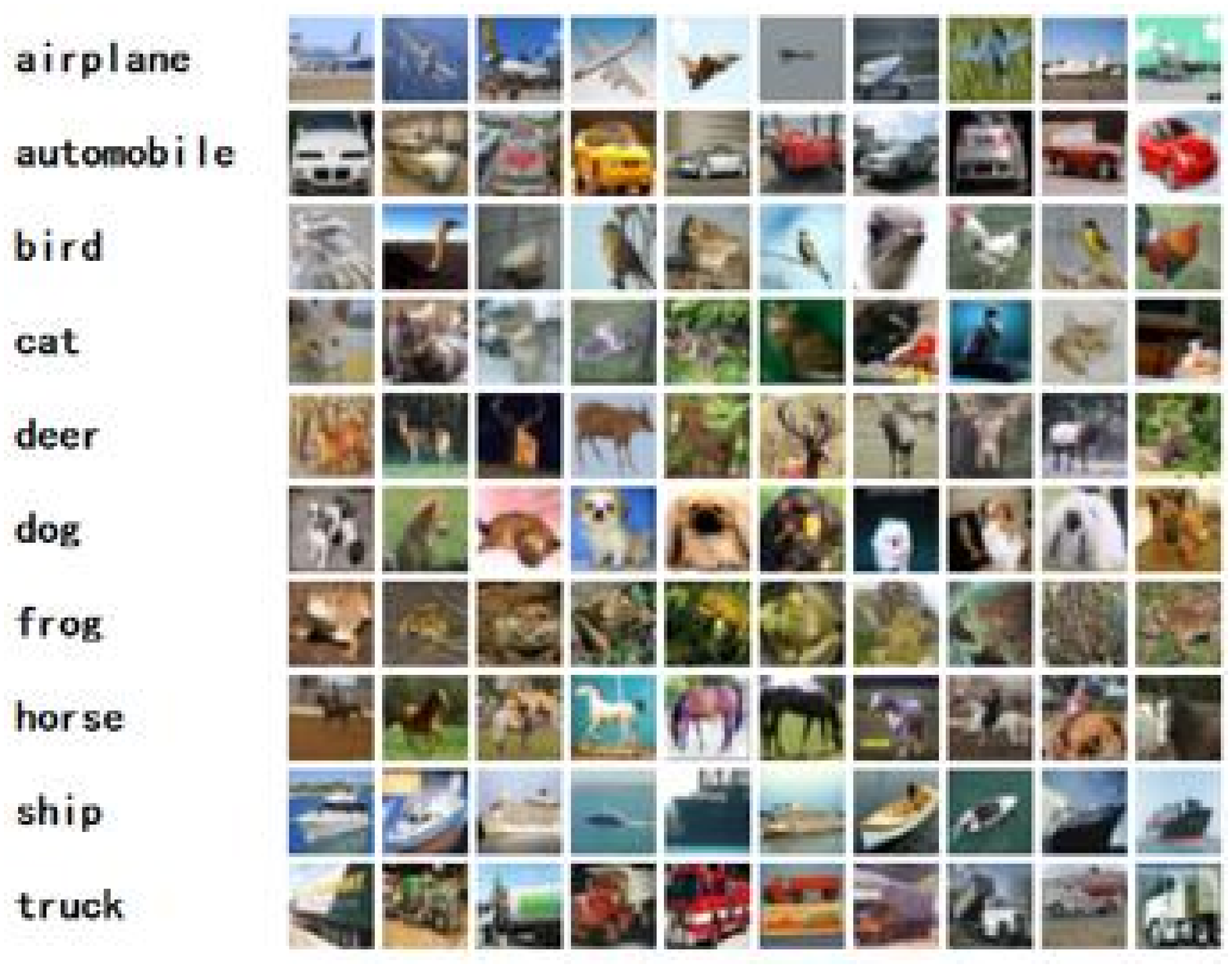}
\caption{CIFAR-10}
\label{fig:7}
\end{figure}

\begin{figure}[!htb]
\centering
\includegraphics[width=0.5\textwidth]{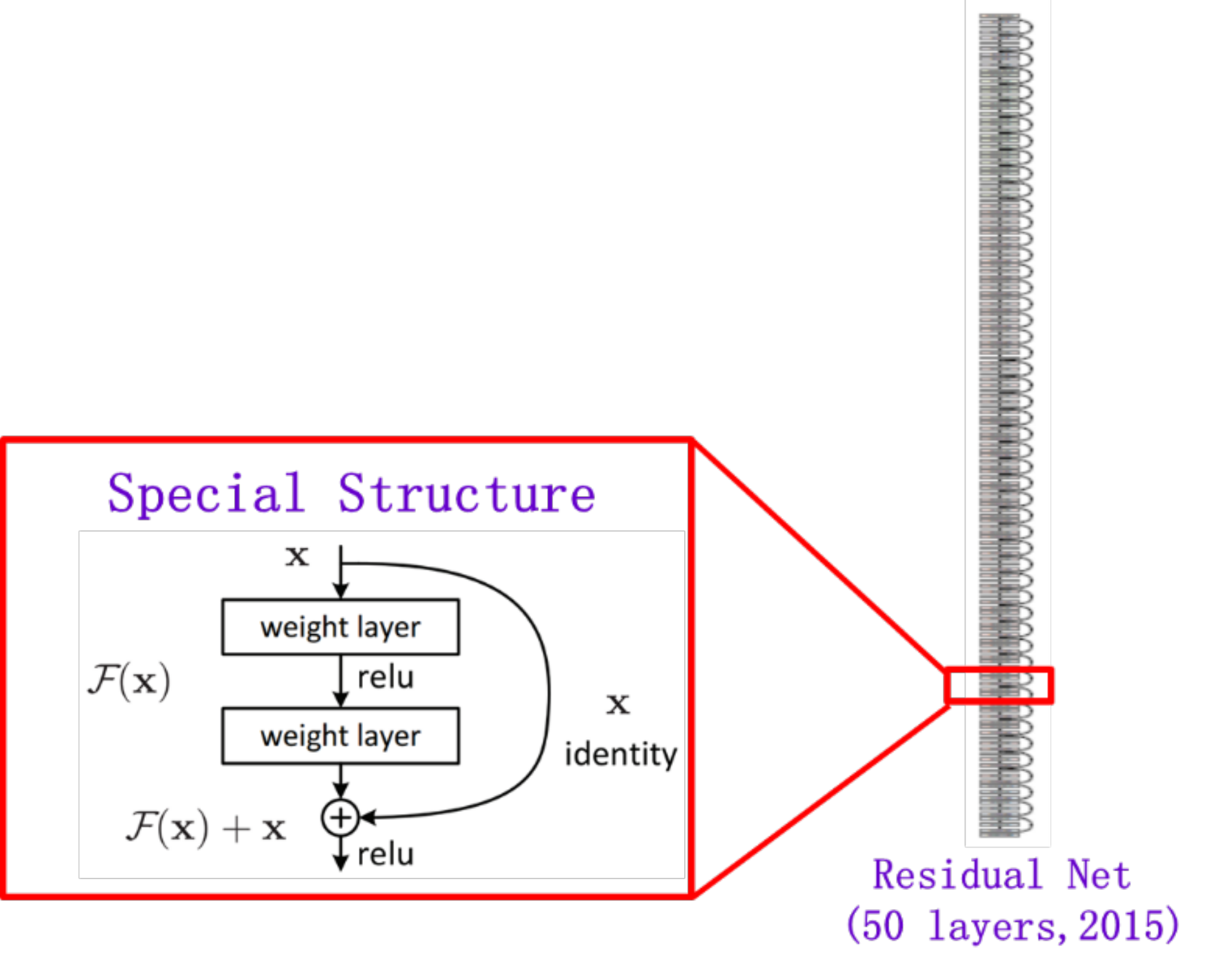}
\caption{The framework of ResNet}
\label{fig:8}
\end{figure}

\begin{table}[!htb]
\centering
\begin{tabular}{c|c|c}
\hline
$--$ & $CIFAR-10$ & $CIFAR-100$ \\
\hline
$common\ ResNet18$ & 94.79\% & 77.06\% \\
\hline
$GO-ResNet18$ & \textbf{95.17\%} & \textbf{77.59\%}  \\
\hline
$common\ ResNet34$ & 95.27\% & 78.26\%  \\
\hline
$GO-ResNet34$ & \textbf{95.77\%} & \textbf{78.72\%}  \\
\hline
$common\ ResNet50$ & 94.44\% & 78.45\% \\
\hline
$GO-ResNet50$ & \textbf{94.72\%} & \textbf{79.50\%} \\
\hline
\end{tabular}
\caption{The model's accuracy rates averaged over five experiments on the test set}
\label{tab:1}
\end{table}

Recognizing objects in an actual scene is not dependent on corresponding domain knowledge but on humans' prior information. For object recognition tasks, the Geometric Operator Convolutional Neural Network's recognition effect is worth exploring. The commonly used public data sets for common object recognition are CIFAR-10 (ten categories, as shown in Fig. \ref{fig:7}) and CIFAR-100 (100 categories). They are all three-channel color images with a resolution of 32$\times$32. The train set contains 50,000 images and the test set contains 10,000 images. As shown in Fig. \ref{fig:8}, ResNet18, ResNet34, and ResNet50 were used on these two public datasets. In the experiment, four paddings were added on the four edges. Then, a random 32$\times$32 cropping was performed, and a data enhancement method was carried out, which involved turning the image up and down. For both testing and training, the images' pixels are normalized to a 0-1 distribution. The Stochastic gradient descent optimization algorithm with 0.9 the momentum \cite{loshchilov2016sgdr} was used during the training process. The batch size was 100, the initial learning rate was 0.1, and the weight decay was 0.0005. The learning rate was reduced by one fifth per 60, 120, and 160 epochs. We report the performance of our algorithm on a test set after 200 epochs based on the average over five runs.

\begin{figure*}[!htb]
\centering
\subfigure[CIFAR-10: ResNet18]{
\label{fig:26}
\includegraphics[width=0.3\textwidth]{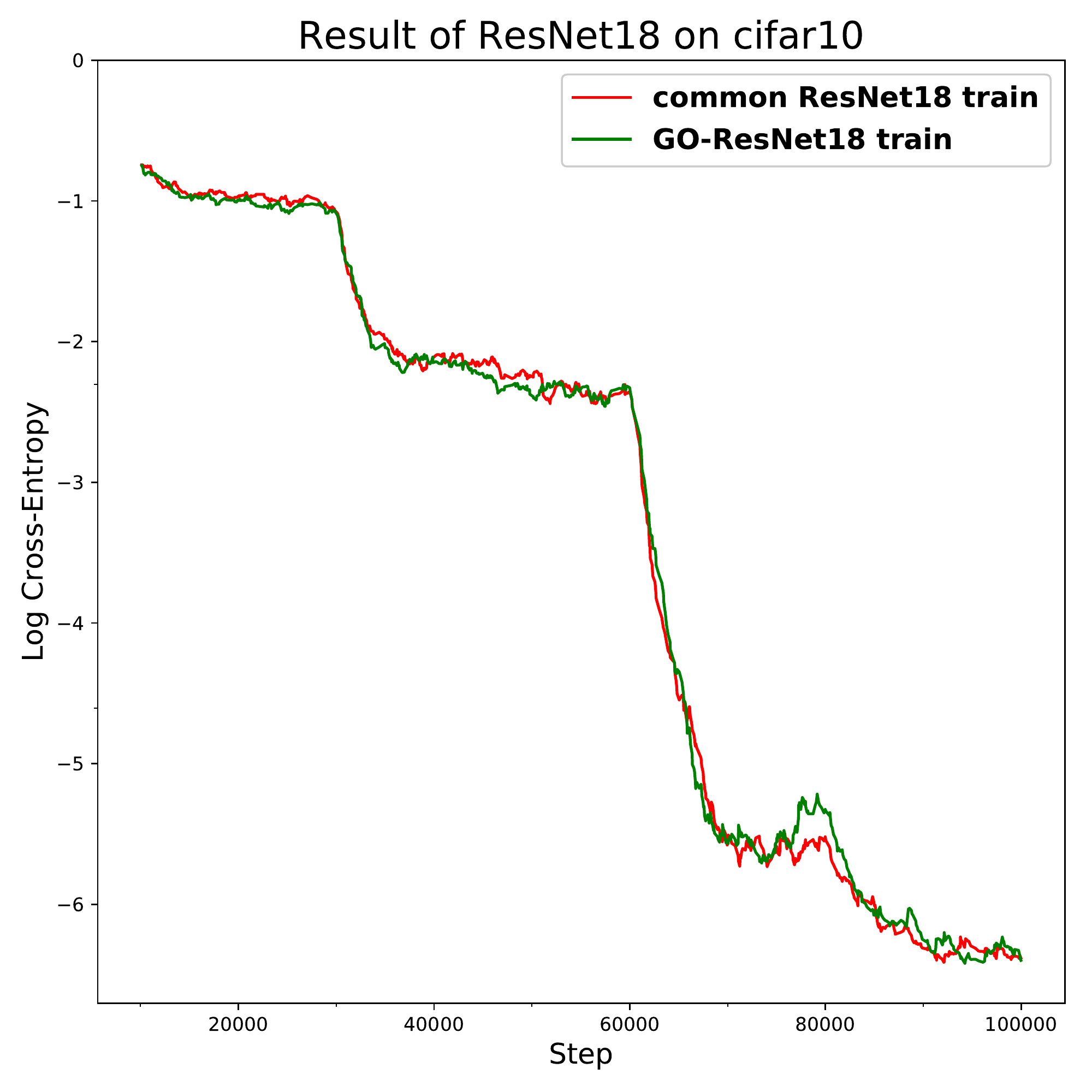}}
\quad
\subfigure[CIFAR-10: ResNet34]{
\label{fig:27}
\includegraphics[width=0.3\textwidth]{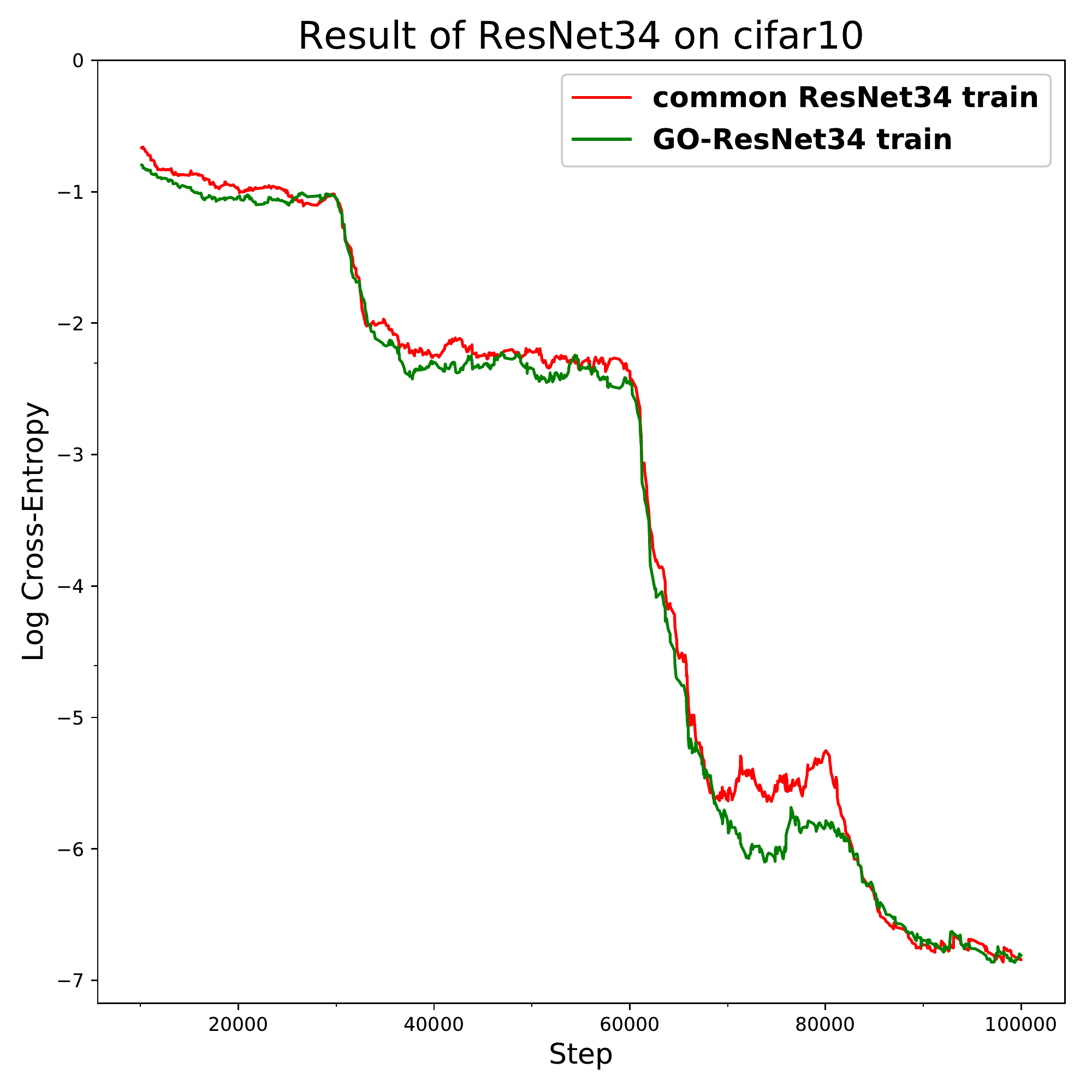}}
\quad
\subfigure[CIFAR-10: ResNet50]{
\label{fig:28}
\includegraphics[width=0.3\textwidth]{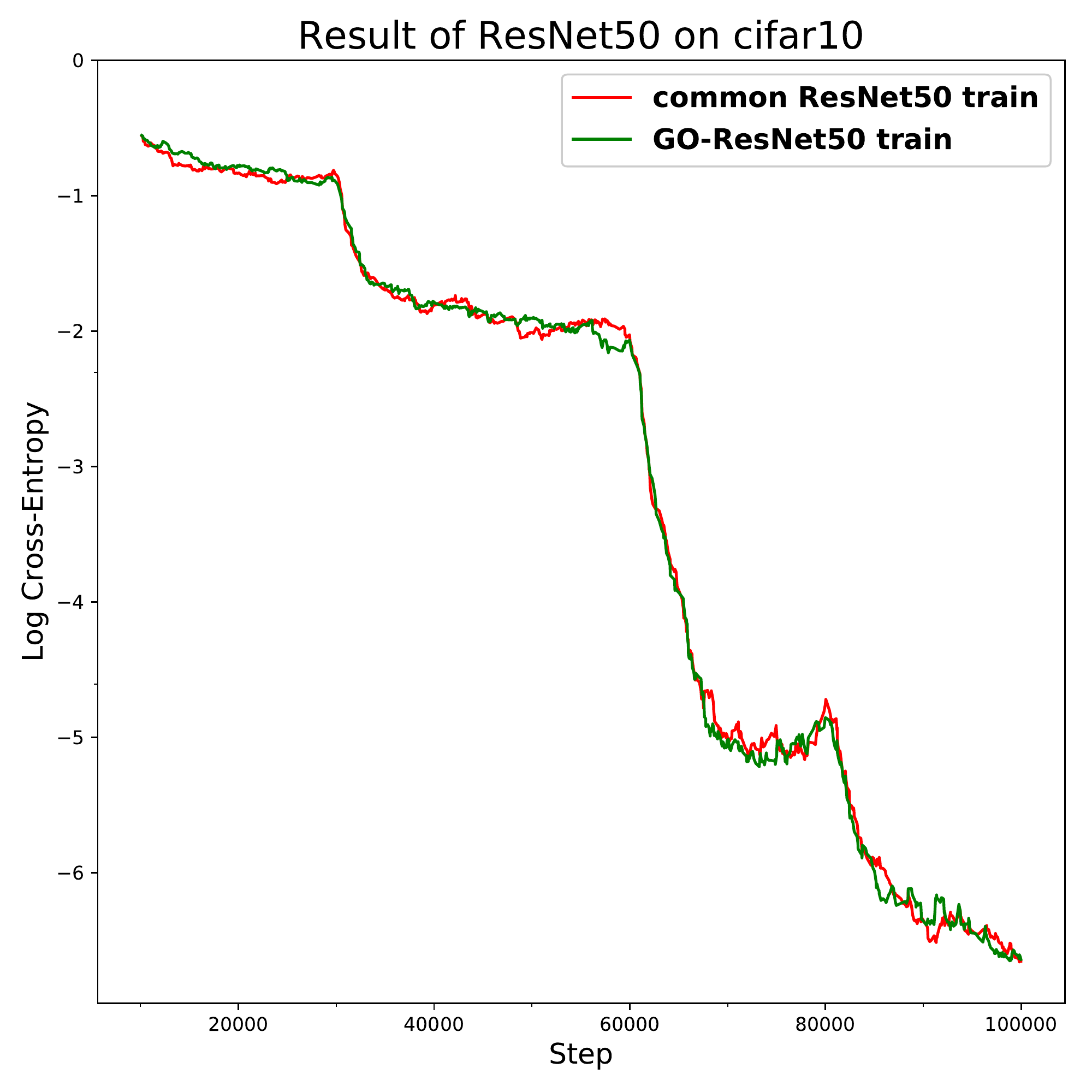}}
\subfigure[CIFAR-100: ResNet18]{
\label{fig:29}
\includegraphics[width=0.3\textwidth]{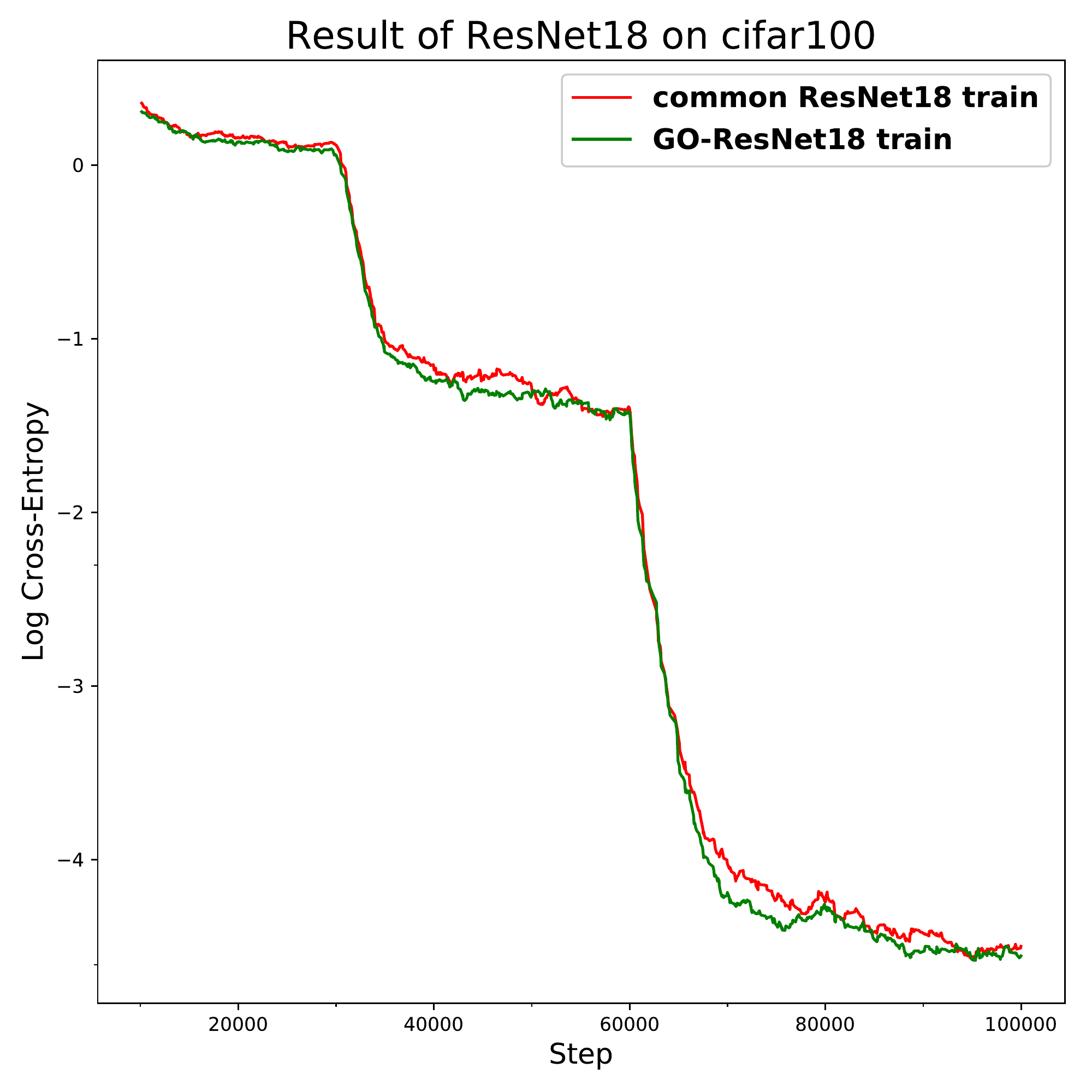}}
\quad
\subfigure[CIFAR-100: ResNet34]{
\label{fig:210}
\includegraphics[width=0.3\textwidth]{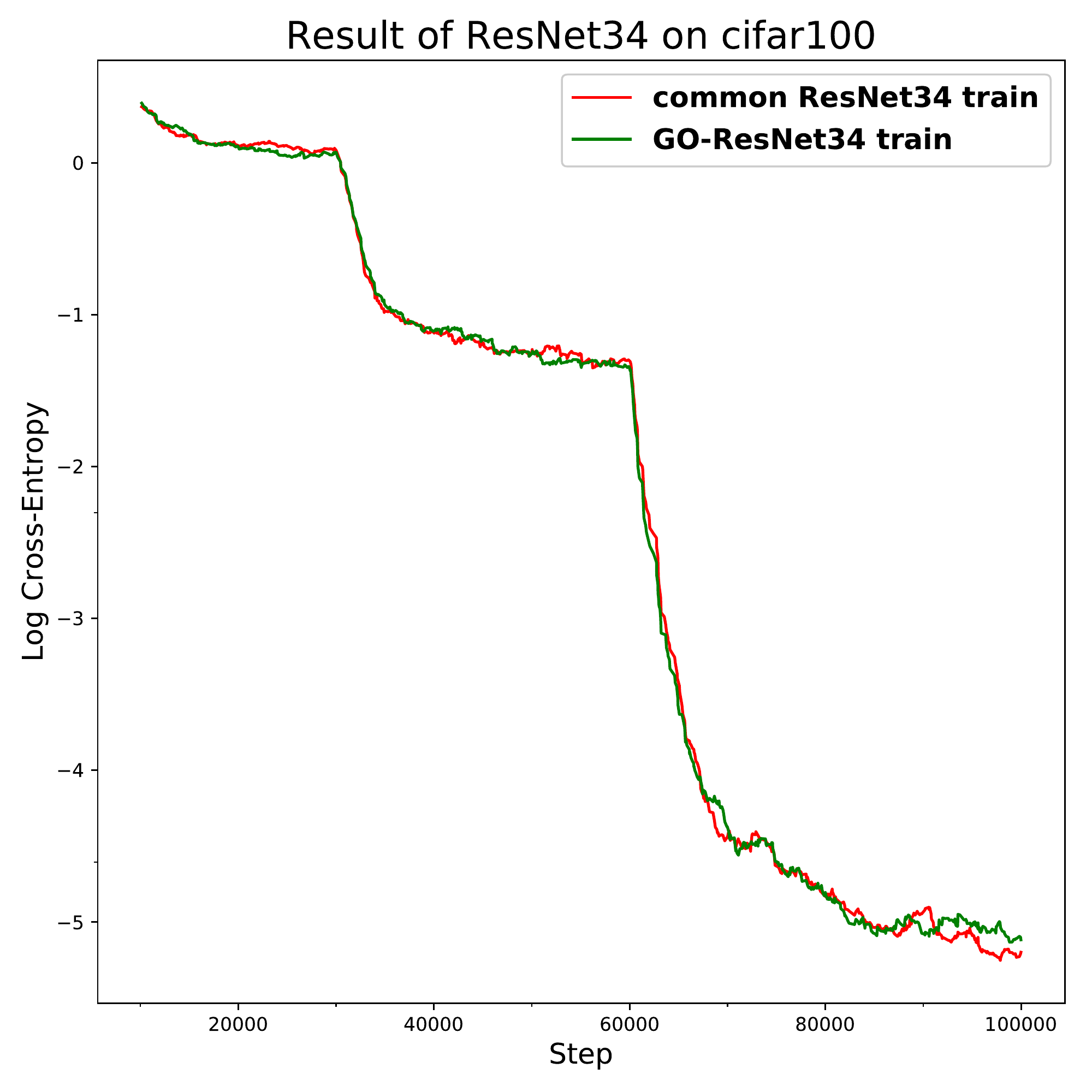}}
\quad
\subfigure[CIFAR-100: ResNet50]{
\label{fig:211}
\includegraphics[width=0.3\textwidth]{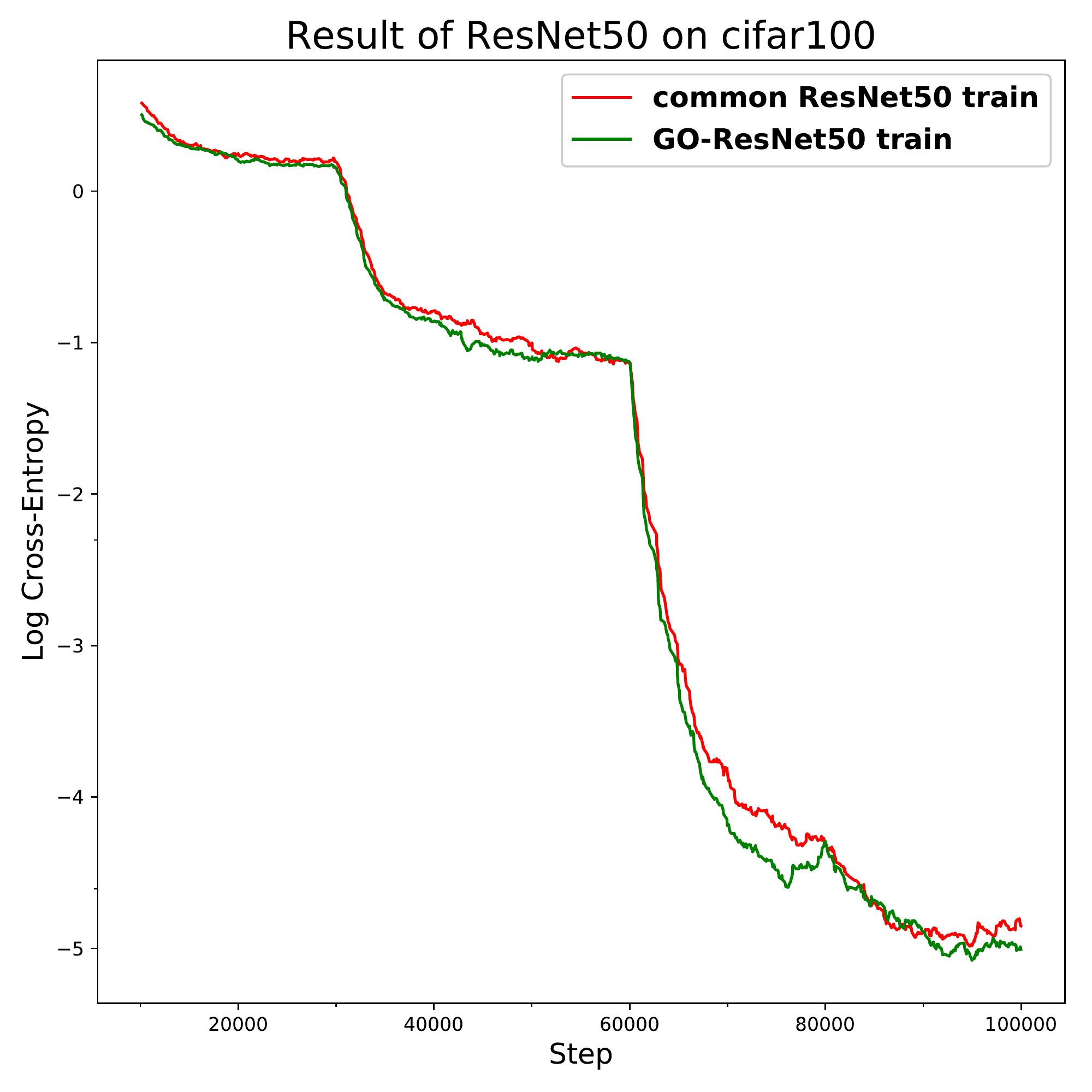}}
\caption{Log of cross entropy curve during training in common ResNet 18-34-50 and GO-ResNet 18-34-50.}
\label{fig:loss-curve}
\end{figure*}

\begin{figure*}[!htb]
\centering
\subfigure[CIFAR-10: ResNet18]{
\label{fig:20}
\includegraphics[width=0.3\textwidth]{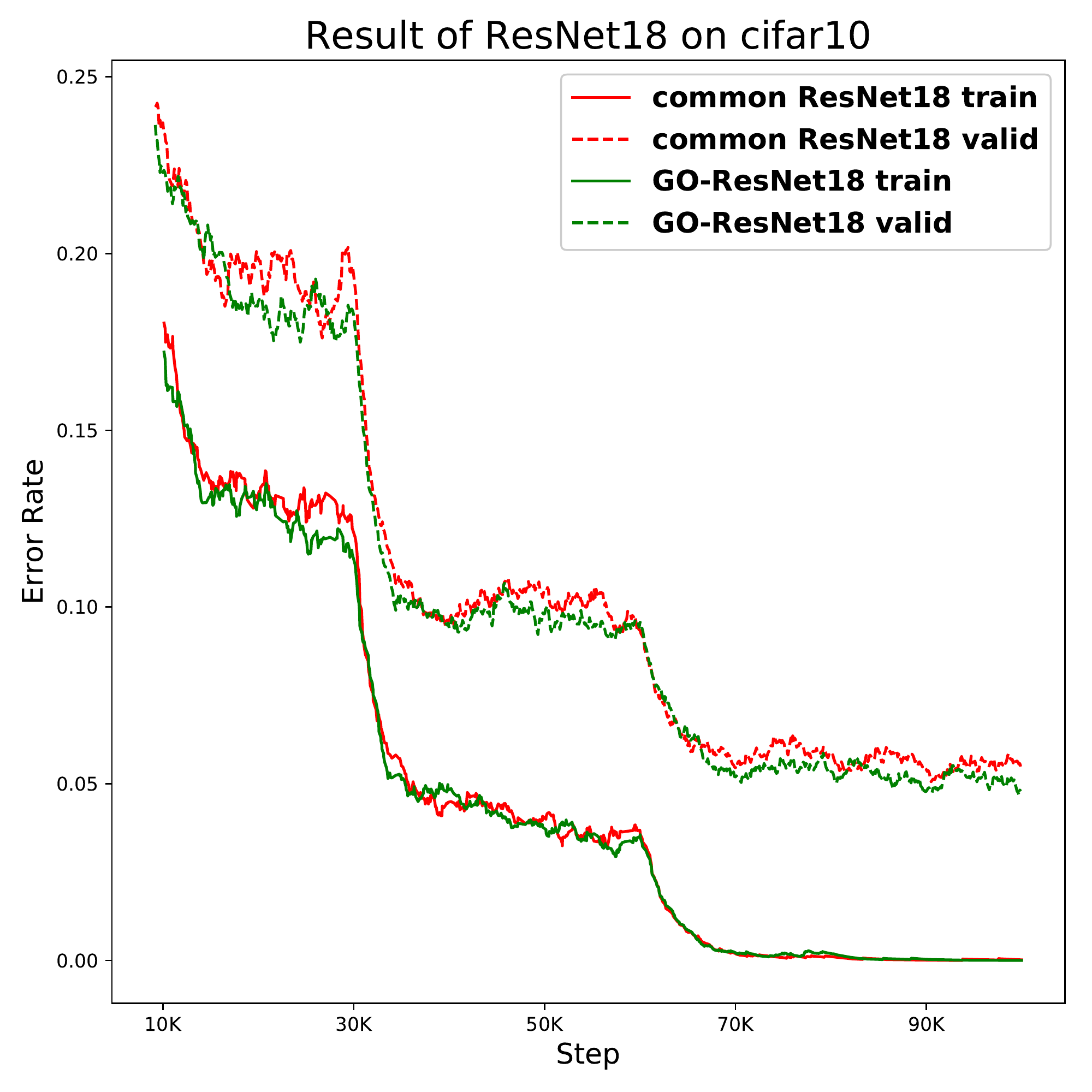}}
\quad
\subfigure[CIFAR-10: ResNet34]{
\label{fig:21}
\includegraphics[width=0.3\textwidth]{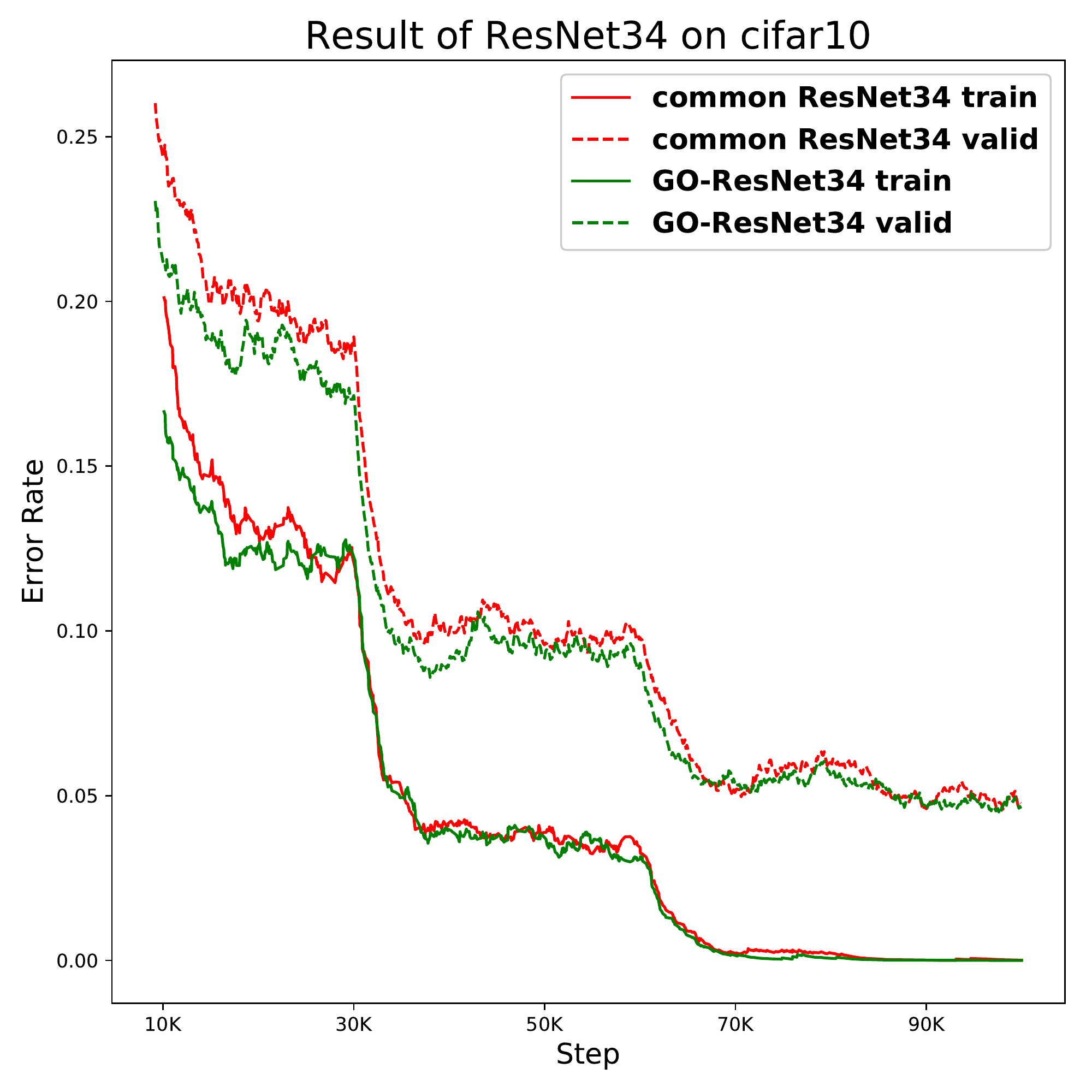}}
\quad
\subfigure[CIFAR-10: ResNet50]{
\label{fig:22}
\includegraphics[width=0.3\textwidth]{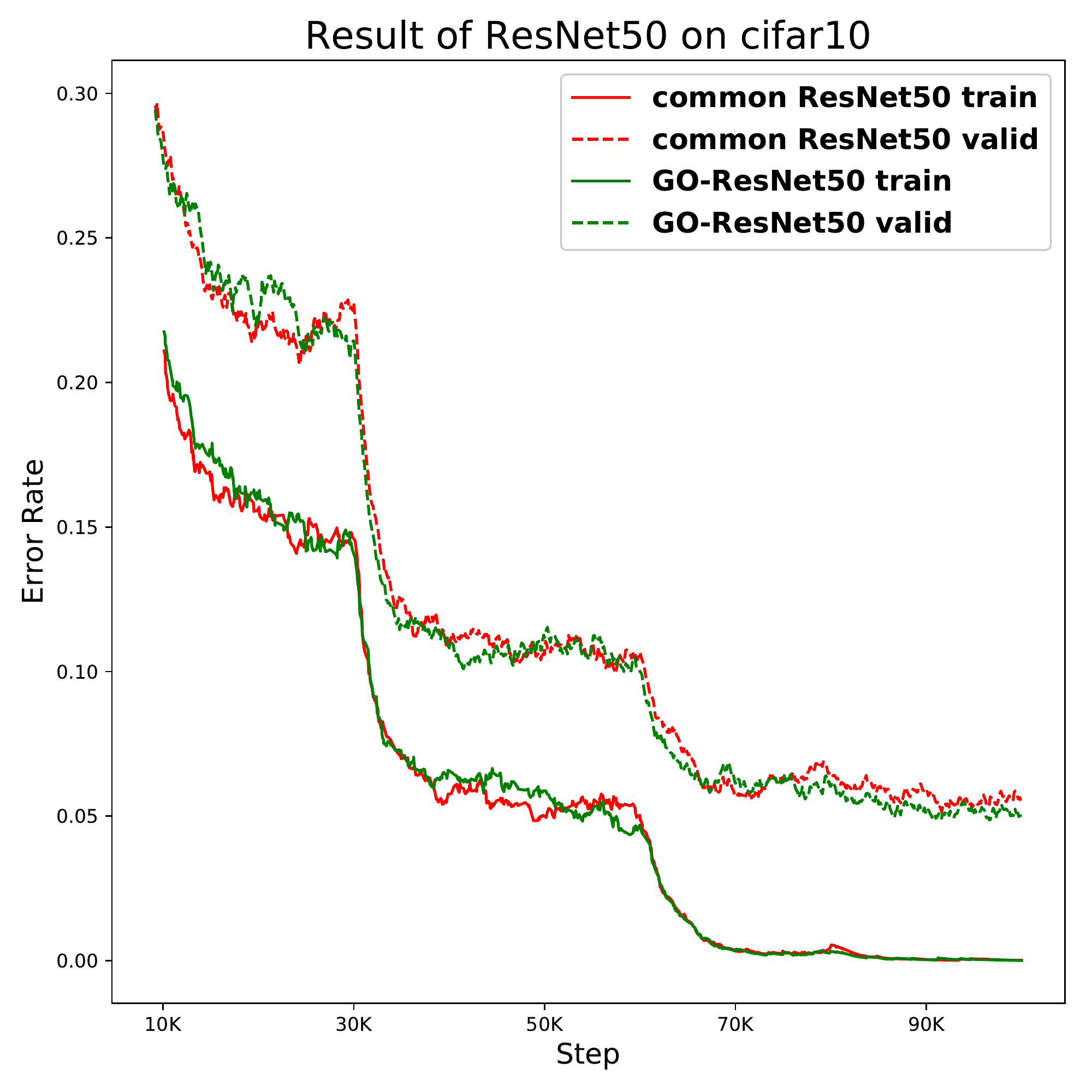}}
\subfigure[CIFAR-100: ResNet18]{
\label{fig:23}
\includegraphics[width=0.3\textwidth]{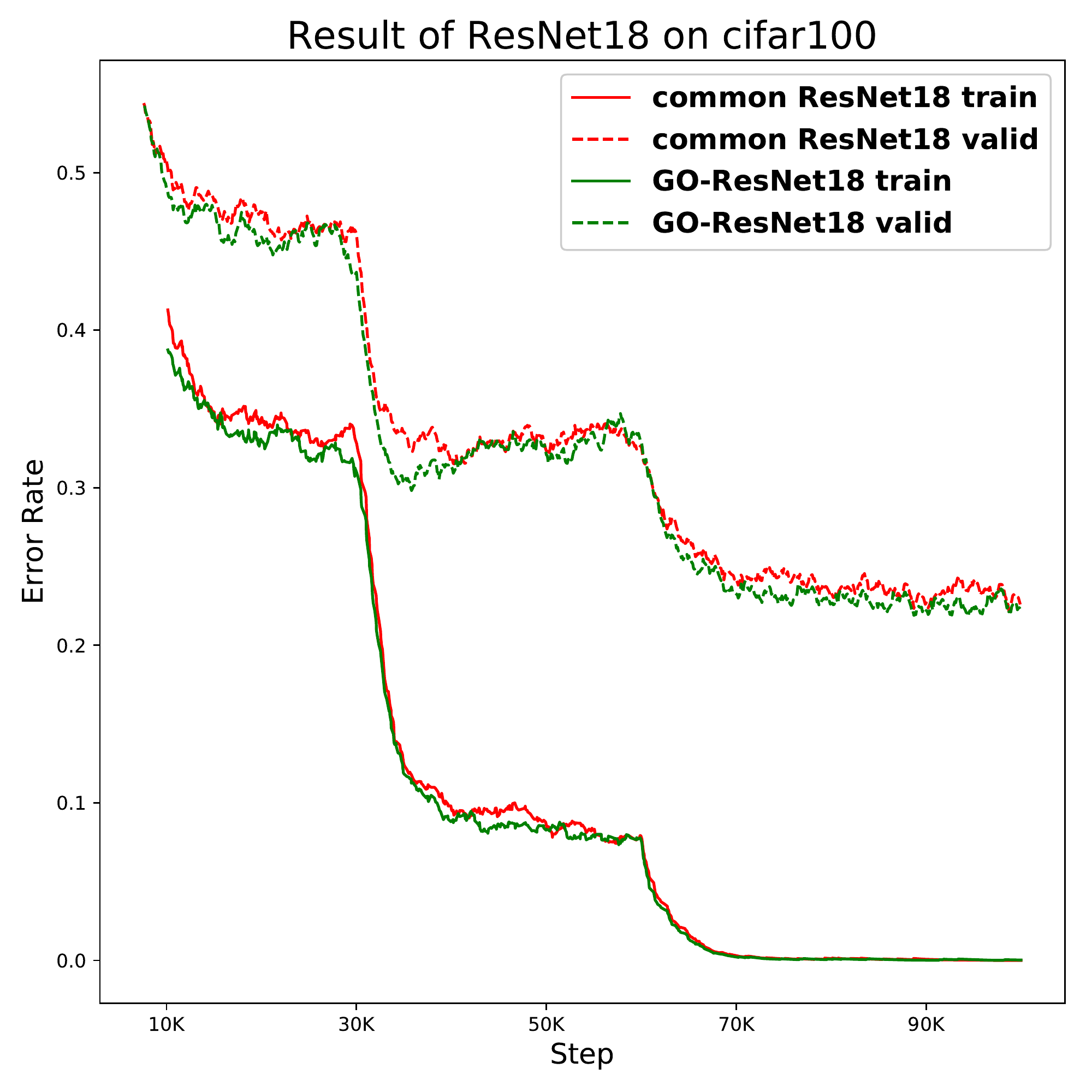}}
\quad
\subfigure[CIFAR-100: ResNet34]{
\label{fig:24}
\includegraphics[width=0.3\textwidth]{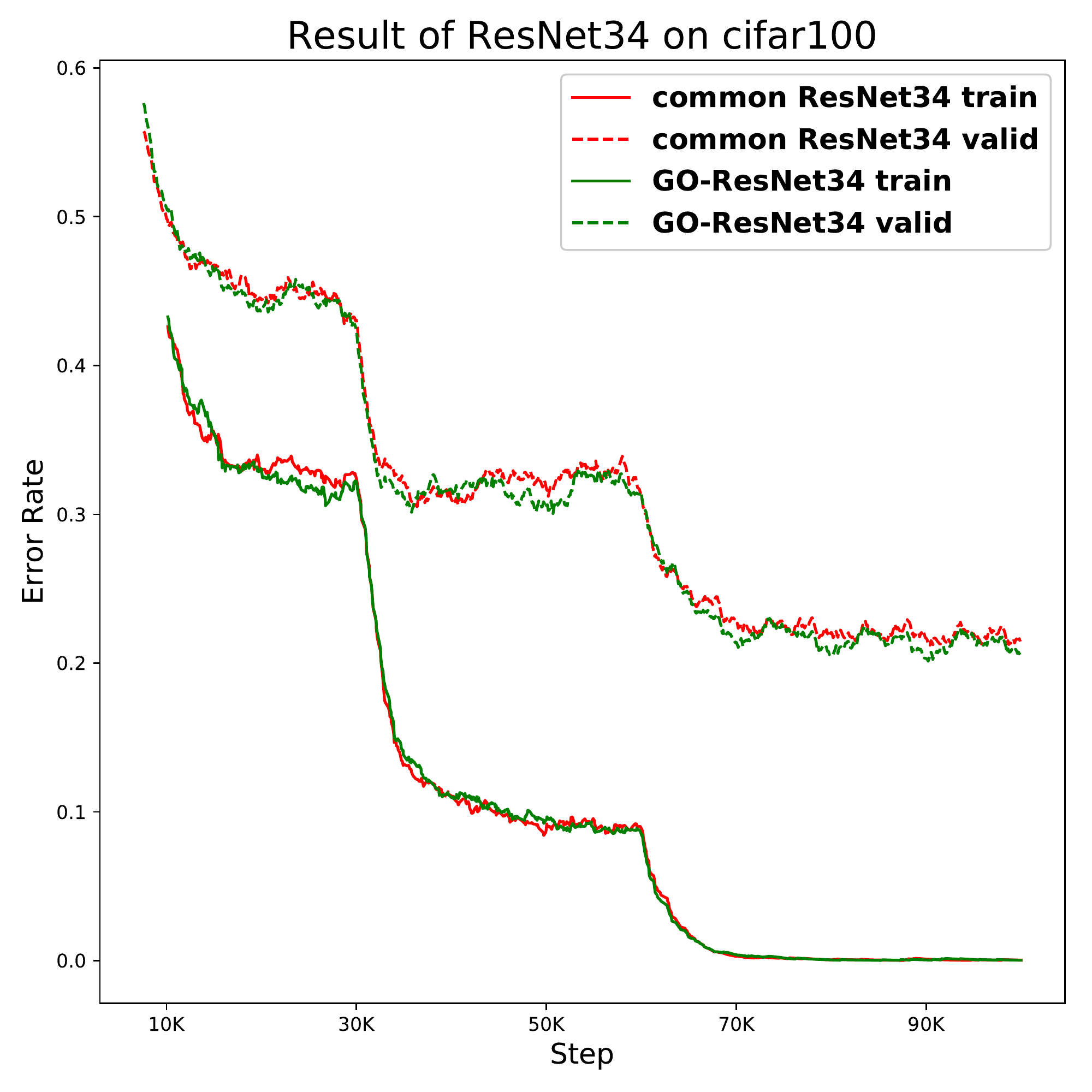}}
\quad
\subfigure[CIFAR-100: ResNet50]{
\label{fig:25}
\includegraphics[width=0.3\textwidth]{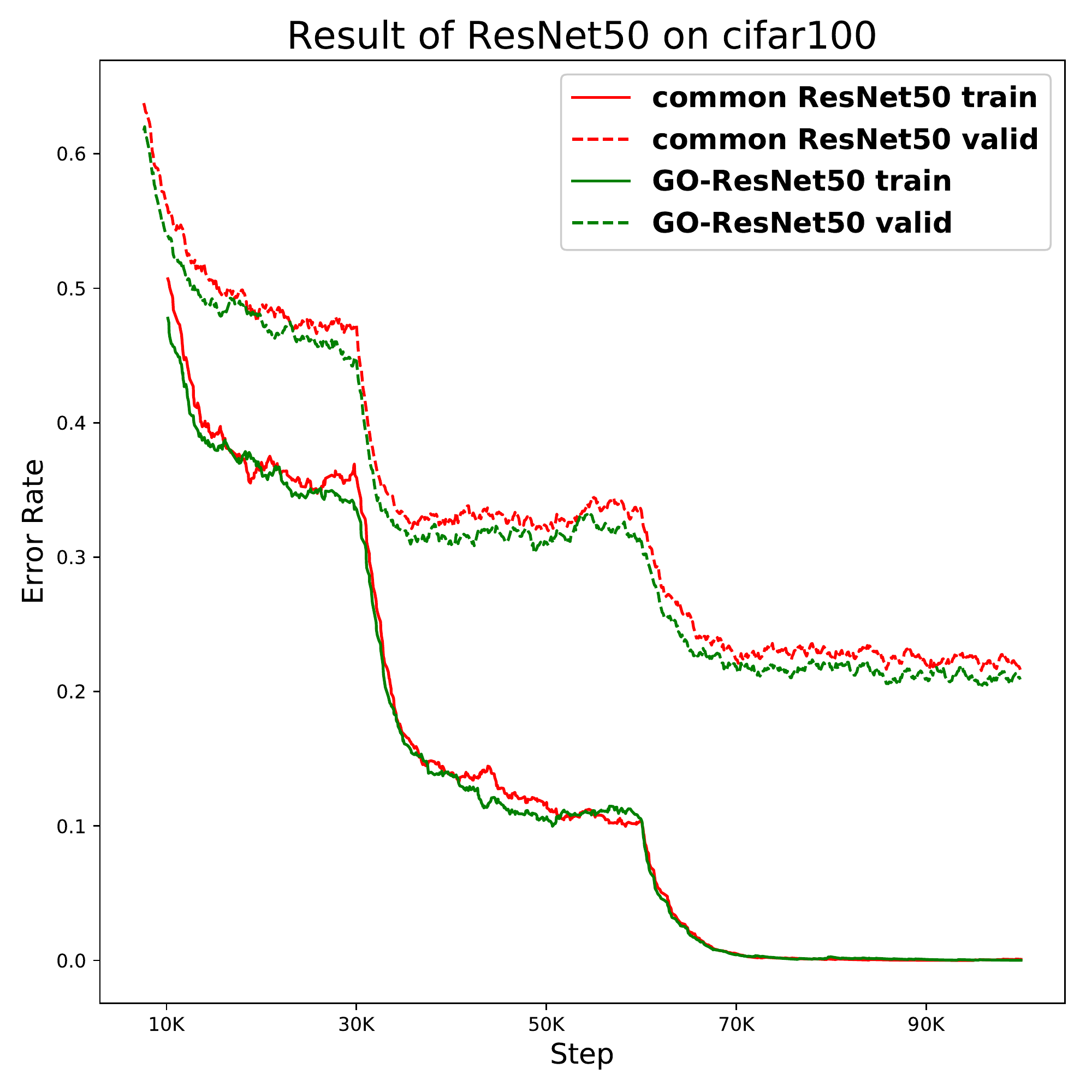}}
\caption{Error rate curve during training in common ResNet 18-34-50 and GO-ResNet 18-34-50.}
\label{fig:acc-curve}
\end{figure*}

As shown in Fig. \ref{fig:loss-curve}, according to the cross-entropy curve of the CIFAR-10 and CIFAR-100 train sets, GO-CNN's value initially fell faster than the common CNN's, eventually almost reaching the same value. It is verified that Geometric Operator Convolutional Neural Network achieves the same approximation accuracy as the common Convolutional Neural Network. According to the error rate curve of the CIFAR-10 and CIFAR-100 verification set (Fig. \ref{fig:acc-curve}), the value of Geometric Operator Convolutional Neural Network is lower than that of the common Convolutional Neural Network. In addition, as shown in Tab. \ref{tab:1}, the Geometric Operator Convolutional Neural Network on the CIFAR-10 test set was 0.4\% more accurate than the common Convolutional Neural Network. On the CIFAR-100 test set, the GO-CNN was 0.5\% more accurate than the common CNN. It is verified that Geometric Operator Convolutional Neural Network achieves the same generalization error bound as the common Convolutional Neural Network.


\textbf{Feature visualization} One way to evaluate a model is through visualizing the features that the model extracts; this is called feature visualization. T-SNE \cite{maaten2008visualizing} or PCA \cite{jolliffe2011principal} are generally used for visualization. The T-SNE visualization maps data points to a two-dimensional or three-dimensional probability distribution through affinitie transformation. Then, the data points are displayed with a two-dimensional or three-dimensional plane.

\begin{figure}[!htb]
\centering
\subfigure[Common CNN]{
\label{fig:101}
\includegraphics[width=0.18\textwidth]{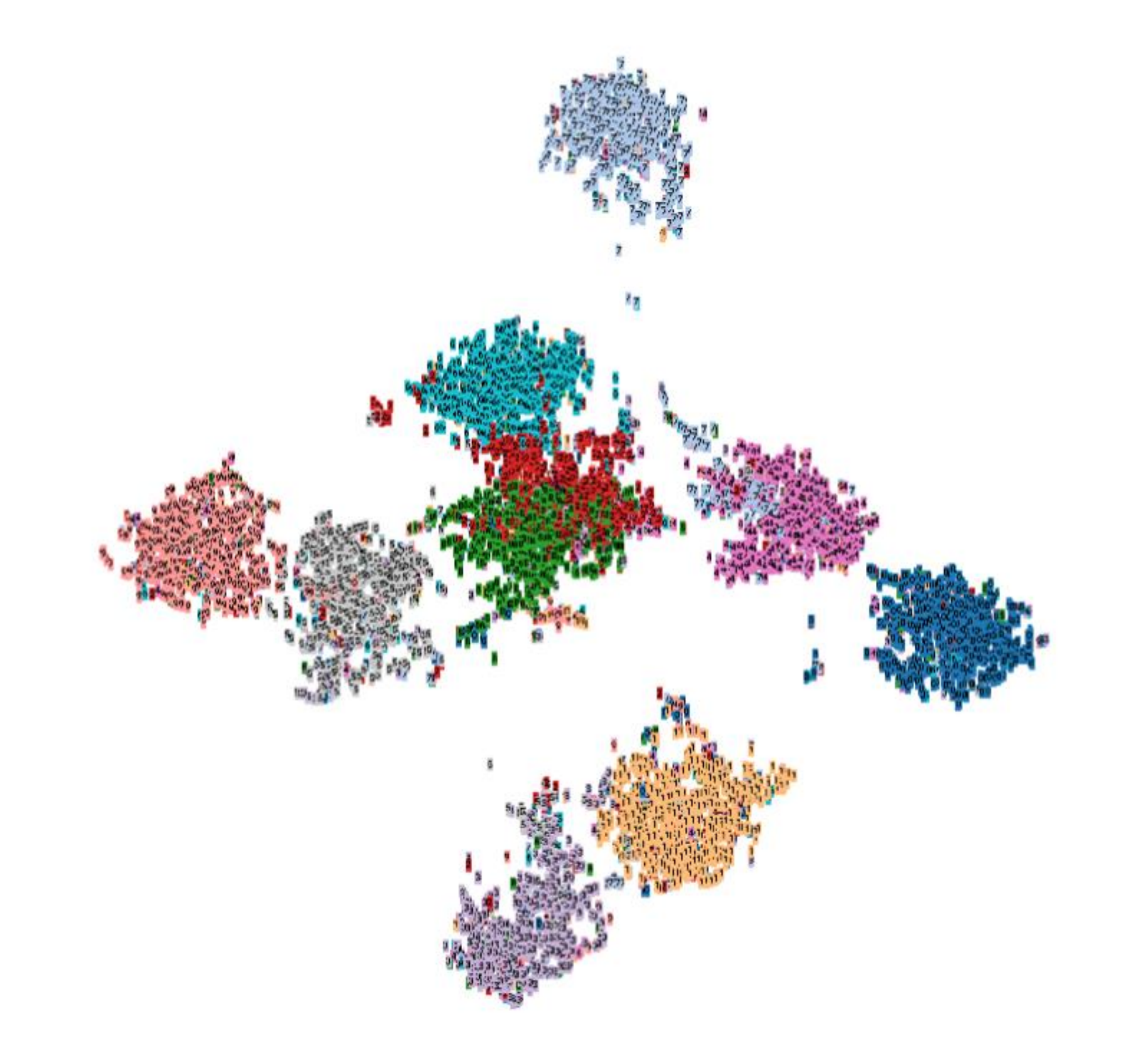}}
\subfigure[GO-CNN]{
\label{fig:102}
\includegraphics[width=0.18\textwidth]{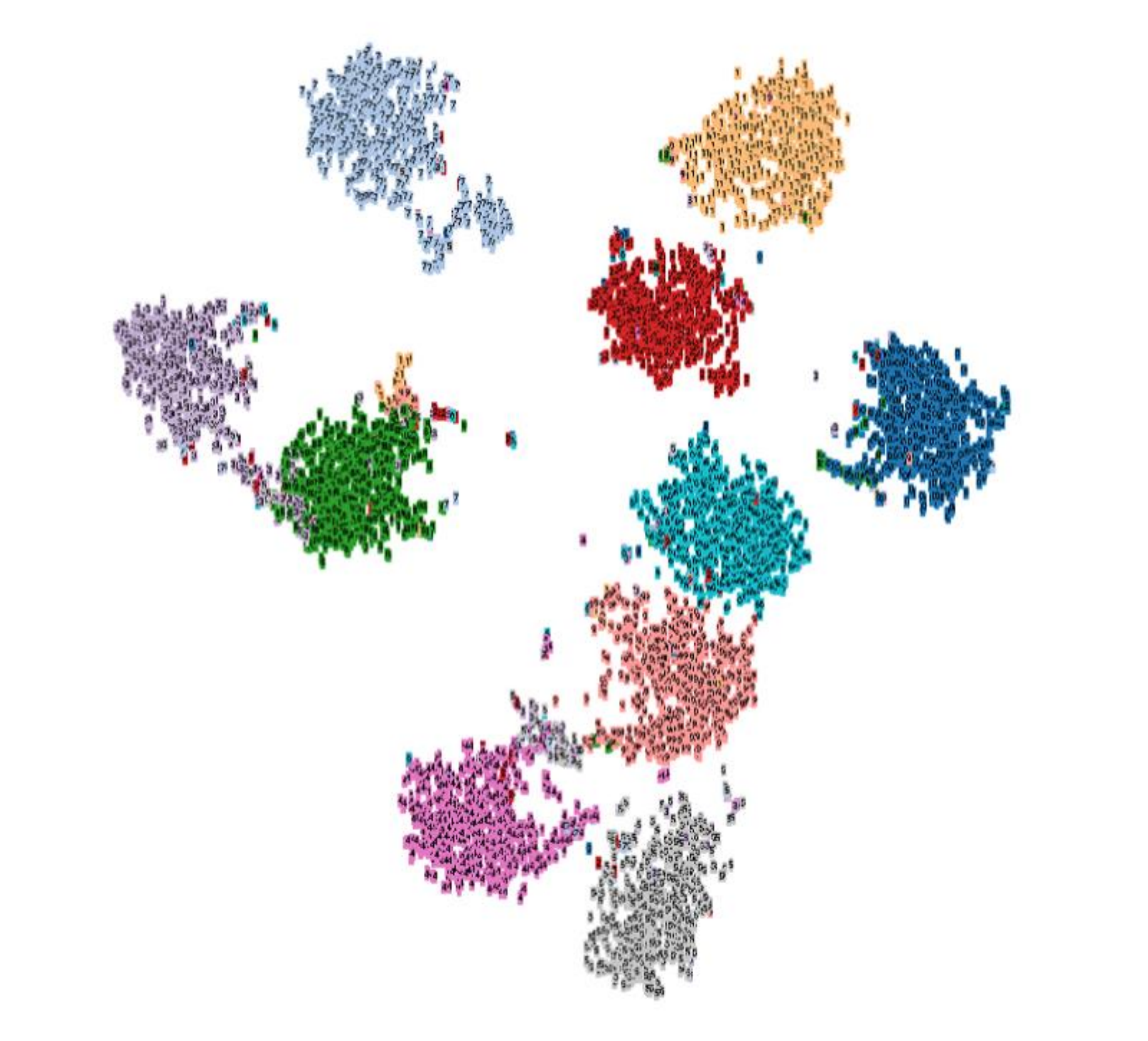}}
\subfigure{
\label{fig:103}
\includegraphics[width=0.08\textwidth]{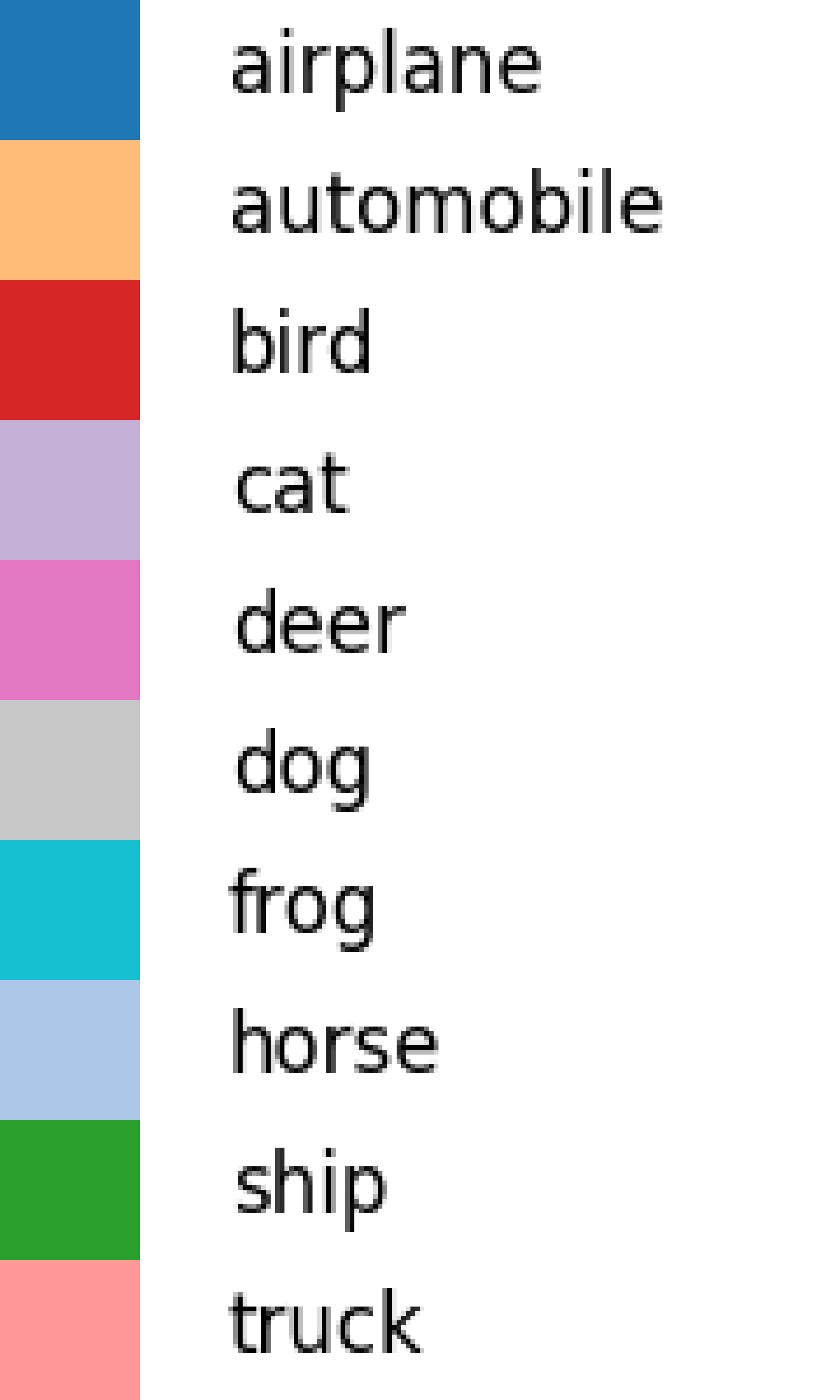}}
\caption{T-SNE two-dimensional visualization of CIFAR-10}
\label{fig:10}
\end{figure}

In this paper, a two-dimensional T-SNE visualization is adopted to display the CIFAR-10 features extracted by the model. As shown in Fig. \ref{fig:10}, the CIFAR-10 features extracted by the Geometric Operator Convolutional Neural Network are evenly separated from each other in the two-dimensional visualization of T-SNE, while the features extracted from the common Convolutional Neural Network are mixed. It is apparent that the features extracted by the Geometric Operator Convolutional Neural Network are more separable; in other words, the features learned by the Geometric Operator Convolutional Neural Network are more distinguishable and easy to classify with the last fully connected layer.

The numerical experimental results and the feature visualizations of the two datasets reveal that the Geometric Operator Convolutional Neural Network achieves the same approximation accuracy and the same upper bound for the generalization error as the common Convolutional Neural Network. Moreover, the features extracted by Geometric Operator Convolutional Neural Network are more distinguishable.

\subsection{Generalization}\label{sec:expe:gener}
In many practical applications, such as the military, medical care, and so on, annotated data are often insufficient. Thus, a model's generalization ability for small data sets is of great importance. The generalization ability refers to the ability of a model to predict unknown data when it has been learned by a certain method.

For the open datasets CIFAR-10/100 and MNIST, their train sets are large and their test sets are small. MNIST is a public, handwritten recognition dataset with a total of ten classes. This dataset is shown in Fig. 10 as a channel image with 28$\times$28 resolution and a clean background. There are 50,000 train sets, 5,000 verification sets, and 10,000 testing sets. In these numerical experiments, the test set is directly used to train the model, and the train set is used to evaluate the model. These experiments assess the generalization ability of the Geometric Operator Convolutional Neural Network and the common Convolutional Neural Network.

Many training techniques have been used in numerical experiments with CIFAR and MNIST. For numerical experiments with the CIFAR-10/100, the techniques and models used are the same as in Sec. \ref{sec:expe:appro}. For numerical experiments with the MNIST data set, the adaptive moment estimation (Adam \cite{kingma2014adam}) optimization algorithm was used. In addition, as an image enhancement strategy, the image padding was increased to 32$\times$32 during the training process. The batch size was set to 11, the initial learning rate was 0.001, and the weight decay was 0.0005. The learning rate stays the same until reaching 20,000 iterations. Consequently, we complete 20,000 iterations on one test set and average the performance over five runs in order to report the final performance evaluation of our algorithm. The basic network structure used in the experiment is LeNet \cite{lecun1998gradient} as shown in Fig. \ref{fig:12}. There are two convolution layers and two fully-connection layers in the network. Similarly, in the Geometric Operator Convolutional Neural Network, the first convolutional layer is replaced by the operator convolutional layer. The convolution kernels from the first layer are composed of trainable Gabor kernels and Schmid kernels. The other convolutional layers are the common convolutional layers.

\begin{figure}[!htb]
\centering
\includegraphics[width=0.5\textwidth]{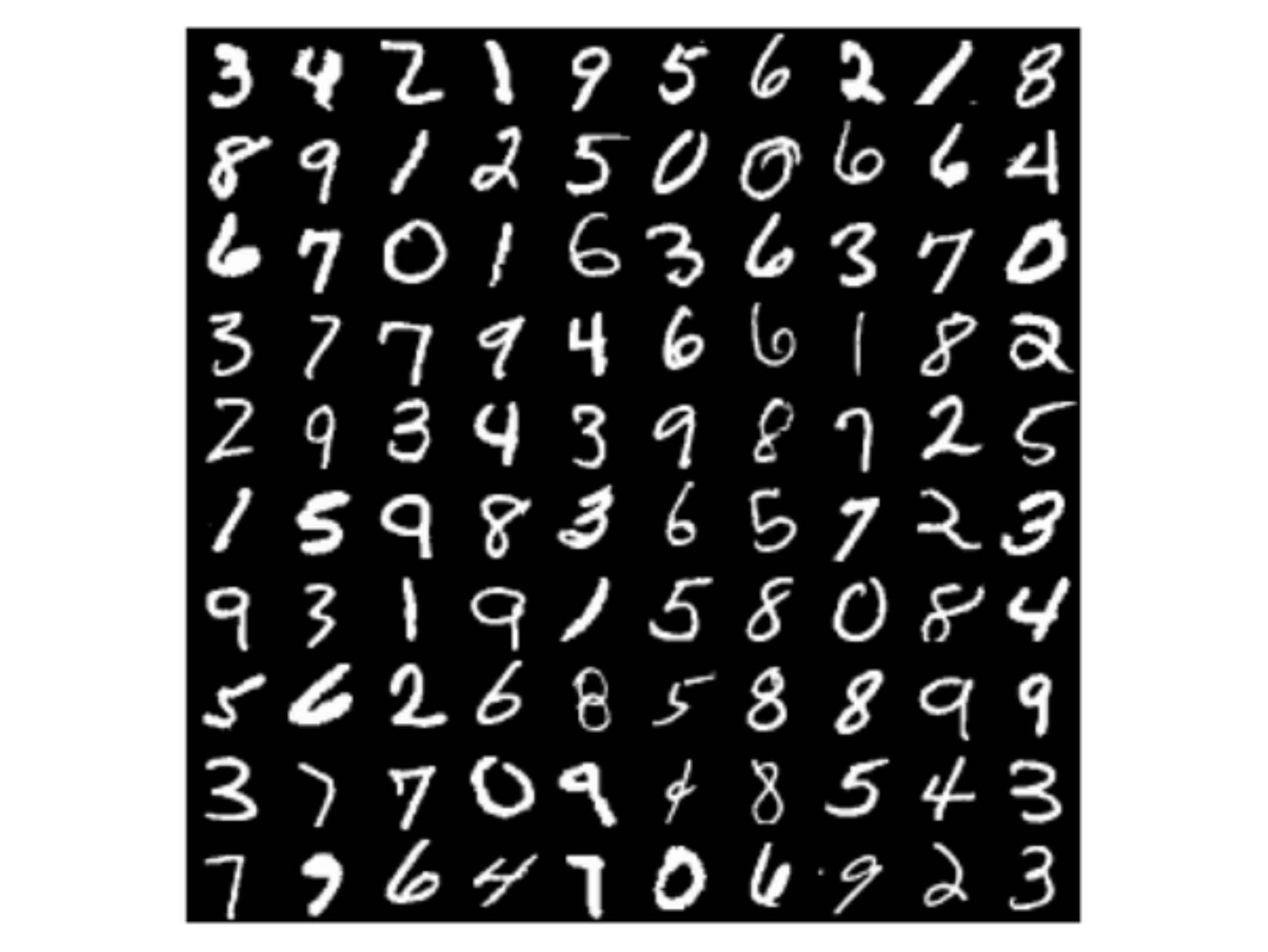}
\caption{MINIST}
\label{fig:11}
\end{figure}

\begin{figure}[!htb]
\centering
\includegraphics[width=0.5\textwidth]{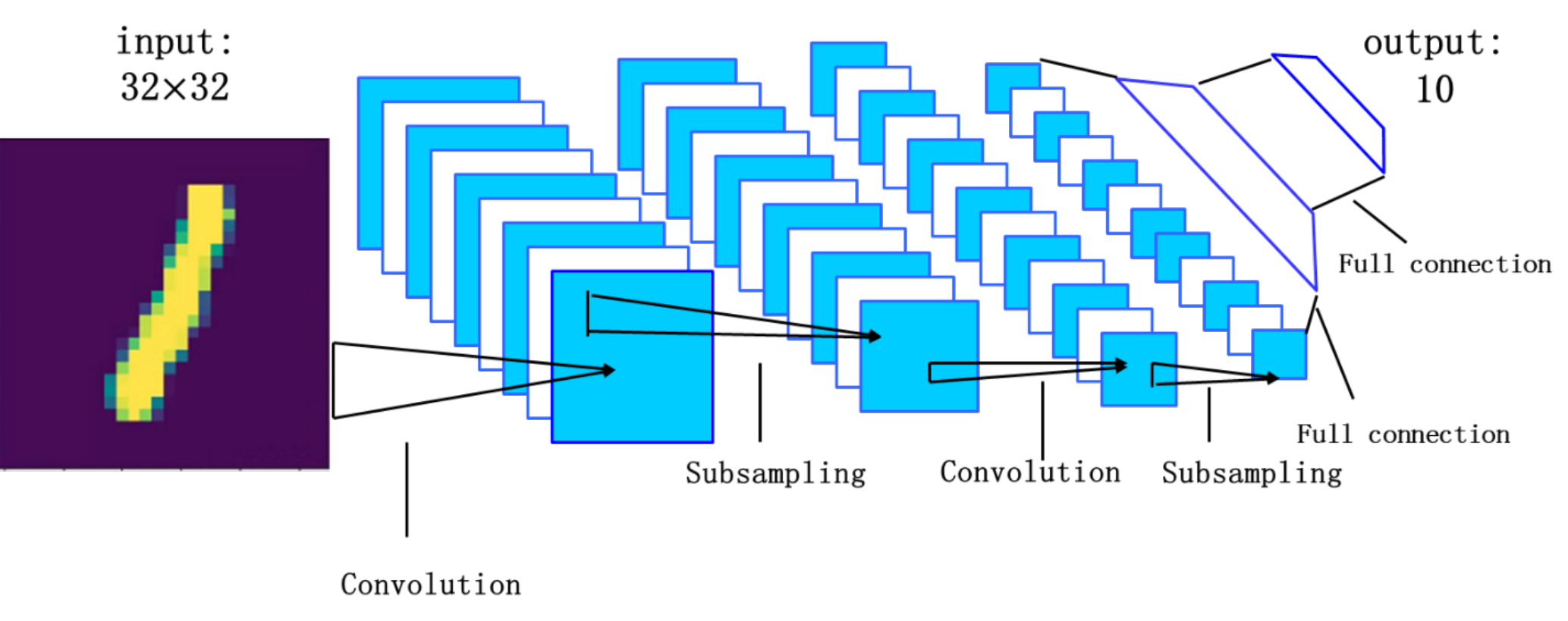}
\caption{The framework of LeNet}
\label{fig:12}
\end{figure}

\begin{table*}[t]
\centering
\begin{tabular}{c|c|c|c}
\hline
$--$ & $CIFAR-10$ & $CIFAR-100$ & $MNIST$ \\
\hline
$common\ ResNet18$ & 84.96\%(94.79\%) & 44.97\%(77.06\%) & --\\
\hline
$GO-ResNet18$ & \textbf{86.21\%(95.17\%)} & \textbf{47.03\%(77.59\%)} & -- \\
\hline
$common\ ResNet34$ & 82.33\%(95.27\%) & 44.74\%(78.26\%)  & --\\
\hline
$GO-ResNet34$ & \textbf{86.36\%(95.77\%)} & \textbf{49.00\%(78.72\%)} & -- \\
\hline
$common\ ResNet50$ & 83.86\%(94.44\%) & 45.93\%(78.45\%) & --\\
\hline
$GO-ResNet50$ & \textbf{85.64\%(94.72\%)} & \textbf{47.09\%(79.50\%)} & -- \\
\hline
$common\ LeNet$ & -- & -- & 97.75\%(99.22\%) \\
\hline
$GO-LeNet$ & -- & -- & \textbf{97.97\%(99.24\%)} \\
\hline
\end{tabular}
\caption{The accuracy of the test sets for the small train set and the large train set (in brackets) as averaged over five experiments}
\label{tab:2}
\end{table*}

As shown in Tab. \ref{tab:2}, from the perspective of the accuracy of MNIST and CIFAR-10/100, after the train set drops to one-fifth of the original train set, the accuracy of the common Convolutional Neural Network falls faster than the Geometric Operator Convolutional Neural Network. Moreover, the Geometric Operator Convolutional Neural Network more accurate than the common Convolutional Neural Network on the original train set. That is to say, the GO-CNN is better at predicting unknown data than the common CNN. The geometric Operator Convolutional Neural Network not only reduces the redundancy of the parameters, but also reduces the dependence on the amount of training samples.

\subsection{Adversarial stability}\label{sec:expe:advers}
Although the Geometric Operator Convolutional Neural Network reduces the number of trainable parameters, it enhances adversarial stability. The current machine learning model, including the neural network and other models, is vulnerable to attacks from adversarial samples. In addition, Convolutional Neural Network shows instability under attacks against adversarial samples \cite{goodfellow2014explaining}. Adversarial samples are produced when an attacker misleads a classifier by slightly disturbing the original sample. It is very important to study the stability of adversarial samples in practice. The false alarm rate of existing intelligent video analysis technology is as much as 30\% to 60\%, which greatly affects the actual application and deployment. For example, the identification system in Tiananmen Square was also removed due to high false alarm rates.

The geometric operator has its own characteristics, and the Schmid operator has rotation invariance. It is worth exploring whether the Geometric Operator Convolutional Neural Network, which is formed by the Schmid operator, enhances the adversarial stability of the adversarial sample when rotated at a certain angle. Geometric operators use domain knowledge and prior knowledge to extract image features. It is worth investigating the Geometric Operator Convolutional Neural Network's ability to enhance the adversarial stability of the adversarial samples against noise interference. The stability of the model is measured by the difference between the accuracy of the original test set and the adversarial sample generated by the test set.


The open handwriting recognition data set (MNIST) is the primary dataset used in this experiment. The techniques and models are the same as those used for MNIST in Sec. \ref{sec:expe:gener}. Both models are trained on the MNIST train set. original images, adversarial samples of gaussian interference, and adversarial samples from random rotation were used to evaluate the two models.

It can be seen from Tab. \ref{tab:3} that when the test set is randomly rotated within 90 degrees, the difference of the Geometric Operator Convolutional Neural Network is 1.21\% lower than that of the common Convolutional Neural Network. This verifies that the Geometric Operator Convolutional Neural Network enhances the adversarial stability of rotated samples. As can be seen from Tab. \ref{tab:4}, when the small Gaussian disturbance (the mean is 0, the standard deviation is 0.3) is applied to the test set, the difference of the Geometric Operator Convolutional Neural Network is 0.6\% lower than that of the common  Convolutional Neural Network. This indicates that the Geometric Operator Convolutional Neural Network enhances the adversarial stability of Gaussian disturbance adversarial samples. In sum, the Geometric Operator Convolutional Neural Network enhances the adversarial stability of certain adversarial samples.

\begin{table*}[t]
\centering
\begin{tabular}{c|c|c}
\hline
$--$ & $common\ LeNet$ & $GO-LeNet$ \\
\hline
$original\ MNIST\ test\ set$ & 99.22\% & \textbf{99.24\%} \\
\hline
$randomly\ rotating\ MNIST\ test\ set$ & 58.97\% & \textbf{60.20\%}  \\
\hline
$difference$ & 40.25\% & \textbf{39.04\%}  \\
\hline
\end{tabular}
\caption{The adversarial stability of rotated samples (the average accuracy over five experiments)}
\label{tab:3}
\end{table*}

\begin{table*}[t]
\centering
\begin{tabular}{c|c|c}
\hline
$--$ & $common\ LeNet$ & $GO-LeNet$ \\
\hline
$original\ MNIST\ test\ set$ & 99.22\% & \textbf{99.24\%} \\
\hline
$Gaussian\ disturbance\ MNIST\ test\ set$ & 95.69\% & \textbf{96.31\%}  \\
\hline
$difference$ & 3.53\% & \textbf{2.93\%}  \\
\hline
\end{tabular}
\caption{The adversarial stability of Gaussian disturbance samples (the average accuracy over five experiments)}
\label{tab:4}
\end{table*}

\subsection{Application}\label{sec:expe:app}
Medical images in China are developing rapidly, but specialist doctors are short of resources, and they are mainly concentrated in big cities and big hospitals. Many small and medium-sized cities do not have sufficient diagnostic imaging capacities, so many patients have to go to big cities in order to access better medical resources and obtain better treatment. Similarly, there are few orthopaedic surgeons in China. Fractures often occur in real life due to accidents, such as falls and car accidents. There are many ways to obtain medical data, such as X-ray images, CT images, MRI images, and representational images; however, orthopedists usually use X-ray images to diagnose fractures. With the development of artificial intelligence technology, many scholars use Convolutional Neural Networks to assist doctors in determining whether a bone image reveals a fracture. \cite{chung2018automated}. 

\begin{figure}[!htb]
\centering
\includegraphics[width=0.5\textwidth]{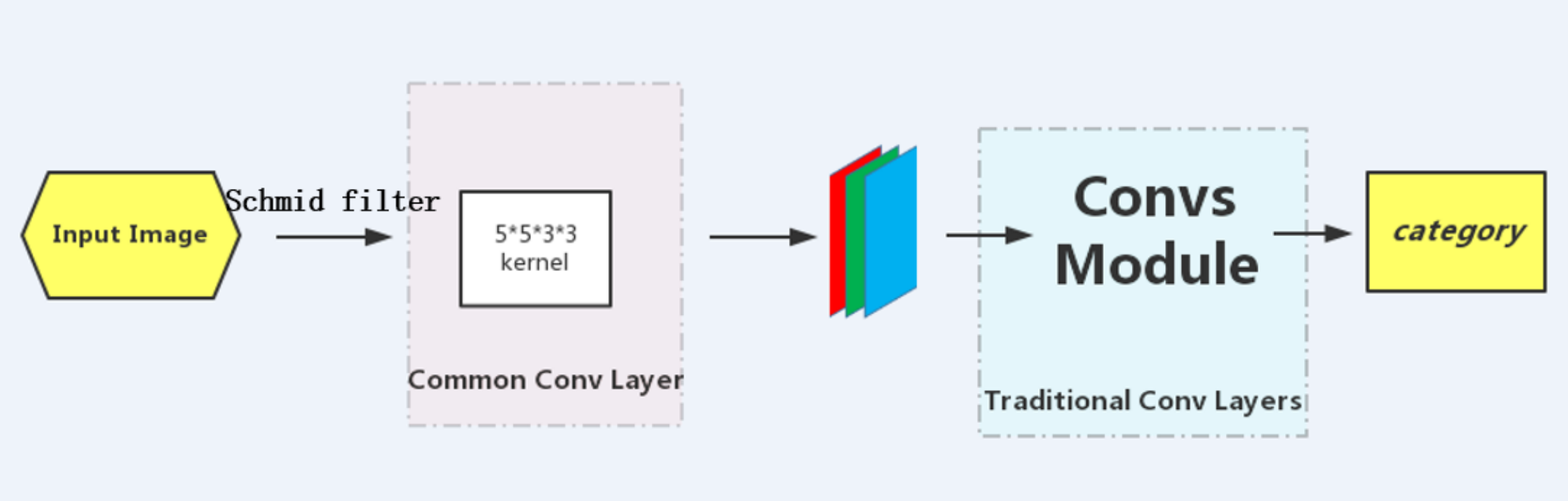}
\caption{The framework of the two-stage method}
\label{fig:13}
\end{figure}

Doctors usually judge whether a fracture has occurred based on whether there is a fracture line (texture) in the image. In \cite{cao2015fracture}, the texture information from the image is used for an auxiliary diagnosis of a fracture. With prior information from the Schmid operator, we do pre-processing by Schmid operators to enhance the texture information from an image. Then, we use the deep Convolutional Neural Network to conduct classification (as shown in Fig. \ref{fig:13}). However, this method, which is preprocessed by geometric operators, can be considered a two-stage method. The parameters of geometric operators are preset by human experience. At this point, it is difficult for the local parameters obtained by the respective optimization to reach the global optimum. Thus, one may consider integrating the preprocessing of geometric operators into the deep network for global parameter learning without prior artificial empirical design parameters. In other words, this would mean using the Geometric Operator Convolution Neural Network proposed in this paper, wherein the convolution kernels from the first layer are all trainable Schmid kernels.

\begin{figure}[!htb]
\centering
\includegraphics[width=0.5\textwidth]{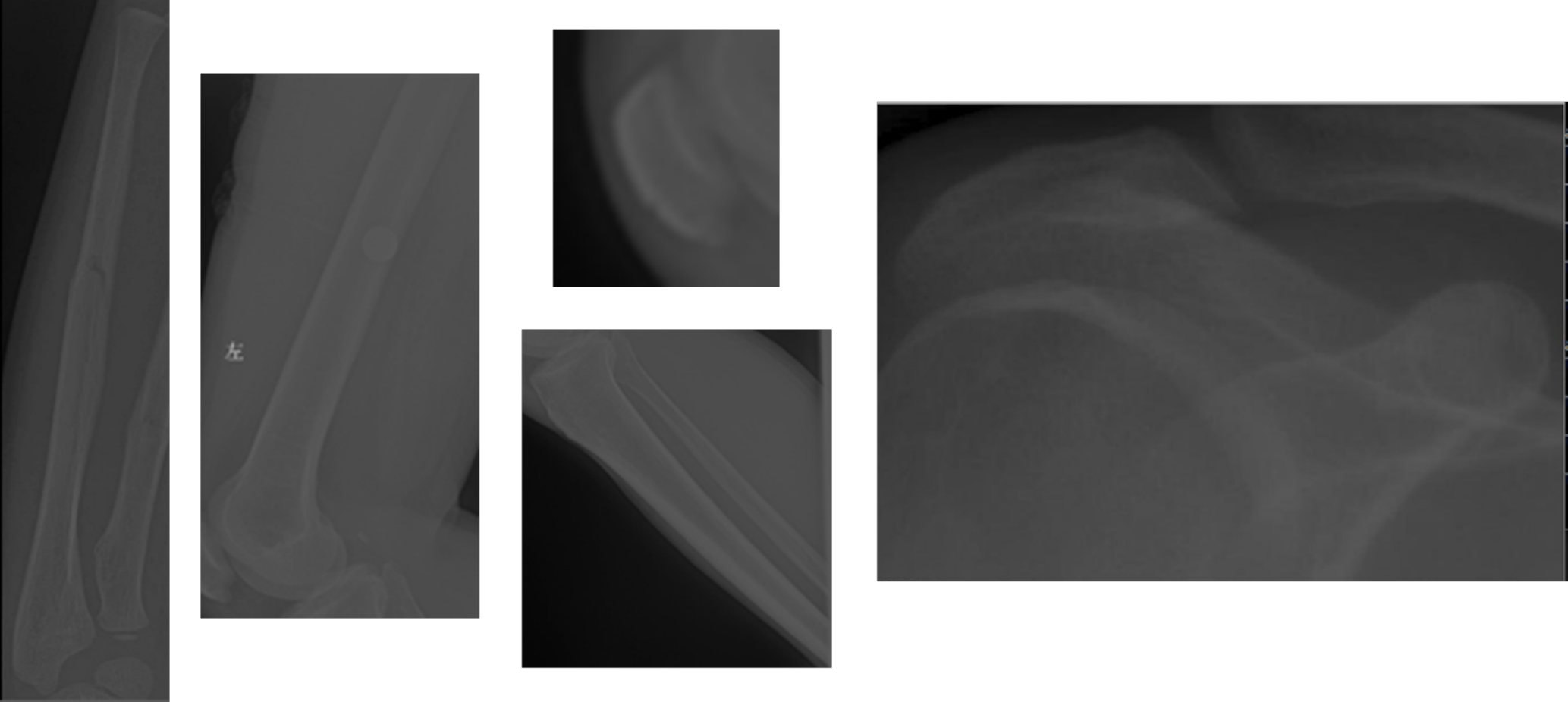}
\caption{Bone images}
\label{fig:14}
\end{figure}

Around 2,000 samples from X-rays taken at the Hainan Peoples Hospital were used as the data for the three kinds of intelligent fracture diagnosis models. Each sample was manually divided into bone regions, as shown in Fig. \ref{fig:14}, with a total of 5,743 bone regions, including 723 bone fracture regions. The above three models are used for numerical experiments. The basic network framework used in the experiment is ResNet50 \cite{he2016deep}, which mainly consists of a new residual structure unit (Fig. \ref{fig:8}). To balance the data during training, the number of fracture patches is increased to 4,016 by rotating the images and changing the background of the images. In the test set, there were 145 fracture patches and 1,004 non-fracture patches. Then, five experiments were conducted to evaluate each model. The stochastic gradient descent optimization algorithm and the finetune strategy were used during the training process, with a batch size of 50. The initial learning rate was 0.001 and the weight decay was 0.0005. The learning rate is reduced by one fifth every 4,000 iterations. Each data class is queued, and the data from each batch is averaged out of each data class during training. We report the performance of our algorithm on the test set after 12,000 iterations based on the average over five runs.

According to Tab. \ref{tab:5}, the Geometric Operator Convolutional Neural Network is the most accurate. Moreover, the fracture recall of the two-stage method is 0.77\% higher than that of the Convolutional Neural Network, indicating that domain knowledge from the field of medicine is important for intelligent diagnosis. The fracture recall of the Geometric Operator Convolutional Neural Network is 2.21\% higher than that of the two-stage method, which indicates that the Geometric Operator Convolutional Neural Network does make use of medical knowledge for fracture diagnosis. The integration of geometric operator into the deep neural network indeed achieve global optimization.

\begin{table*}[!htb]
\centering
\begin{tabular}{c|c|c|c}
\hline
$the\ test\ set$ & $CNN$ & $two-stage\ method$ & $GO-CNN$\\
\hline
$accuracy$ & 92.38\% & 93.05\% & \textbf{93.98\%}\\
\hline
$fracture\ recall$ & 87.97\% & 88.74\% &\textbf{90.95\%}  \\
\hline
$non-fracture\ recall$ & 96.57\% & 96.17\% & \textbf{96.87\%} \\
\hline
\end{tabular}
\caption{Experimental results of intelligent diagnosis}
\label{tab:5}
\end{table*}

In the above experiments, the Geometric Operator Convolutional Neural Network uses a priori knowledge from the field of medicine and provides a better recognition effect. Although the trainable parameters decrease, GO-CNN still reaches the same approximation accuracy and a slightly lower generalization error upper bound when compared with the common CNN. The features extracted from the Geometric Operator Convolutional Neural Network are more distinguishable, and the Geometric Operator Convolutional Neural Network reduces the dependence on training samples and enhances the adversarial stability of certain adversarial samples. Moreover, the GO-CNN also uses medical knowledge for the practical purpose of assisting in intelligent medical diagnoses of bone fractures.


\section{Conclusion and Future Research}\label{sec:con}
In this paper, we present a new framework named the Geometric Operator Convolution Neural Network, where the kernel in the first convolutional layer is replaced with kernels generated by geometric operator functions. This new network boasts several contributions. Firstly, the Geometric Operator Convolution Neural Network is customizable for diverse situations. Other geometric operators may be convolved using the convolution process of the convolution of the Gabor operator and the Schmid operator. Whereas the geometric operator convolution in the GO-CNN can be replaced by different geometric operators, so the GO-CNN is highly versatile. Second, there is a theoretical guarantee in the learning framework of the Geometric Operator Convolutional Neural Network. In this paper, the universal approximation theorem and multiple lemmas are used to prove that GO-CNN reaches the same approximation accuracy and the same generalization error upper bound as the common CNN when certain conditions (i.e., when the training sample is singular) are satisfied. In addtion, through experiments on CIFAR-10/100, we verify that GO-CNN reaches the same approximation accuracy with a smaller generalization error when compared to the common Convolutional Neural Network. Thirdly, the Geometric Operator Convolutional Neural Network reduces the dependence on training samples. The train set and the test set of CIFAR-10/100 and MNIST were exchanged and then re-trained and tested. The Geometric Operator Convolutional Neural Network improved the generalization performance by achieving a higher testing accuracy with the same training loss. In other words, GO-CNN has less dependence on the training samples. Lastly, the Geometric Operator Convolutional Neural Network enhances adversarial stability. Gaussian perturbation and random rotation were performed on the MINIST test set and then tested. The experimental results show that the Geometric Operator Convolutional Neural Network enhances adversarial stability. Furthermore, the GO-CNN improves diagnostic efficiency by offering intelligent medical diagnostic assistance based on domain knowledge acquired from images of bone fractures.

In this paper, only the convolutions of two kinds of geometric operators are considered. In the future, we can explore more submodules suitable for the Geometric Operator Convolutional Neural Network, namely, the convolutions of more geometric operators with better performances. We can explore a more appropriate geometric operator convolution block. In addition, we can analyze the internal relations of the Geometric Operator Convolution Network from the theoretical analysis provided in this paper. 

\section{Acknowledgments}\label{sec:acknowledge}

We thank Professor Zhouwang Yang for his technical guidance. Notably, we would like to express our full thanks to Shiwei Wang and Haikou People's Hospital for providing medical data.

\appendix

\section{Appendix}

%
%
%

\begin{proof}[\textbf{Proof of Proposition1:}]
~\\
Assume that the proposition is not true, then there exist $I_1 \neq I_2$, such that $I_1 * w = I_2 * w$. Thus, if we set $I = I_1 - I_2$, we have $I * w = (I_1 - I_2) * w = 0$, since $*$ is a linear operator, which means that $I = 0$ according to the condition. Therefore, the assumption is not true, and the conclusion is proved.
\end{proof}


\begin{proof}[\textbf{Proof of Proposition2:}]
~\\
Assume that there exists $I \in \mathbb{R}^{3 \times 3}, I \neq 0$, such that $I * k = 0$ holds for $\forall k \in \mathcal{K}_f$.

\noindent We write $I$ in the following matrix way:

\begin{equation}
I = \begin{bmatrix}
a_{00}  &  a_{01}  &  a_{02} \\
a_{10}  &  a_{11}  &  a_{12} \\
a_{20}  &  a_{21}  &  a_{22} \\
\end{bmatrix}
\tag{14.1}
\label{prop2:eqn1}
\end{equation}

We define the \textit{pixel generator function} $f_{ij}$ to be $f_{x-1, y-1}(\theta, \sigma, \gamma, \lambda, \psi)$. Then, we have the following equivalence:

\begin{equation}
I * k = 0 \iff \sum_{i=0}^2 \sum_{j=0}^2 a_{ij}f_{ij} = 0.
\tag{14.2}
\label{prop2:eqn2}
\end{equation}

We will choose a variety of different parameters to discuss.

~\\

\noindent \textbf{\RomanNumeralCaps{1}}\label{prop2:sit1}. $\theta=0,\ \lambda=1,\ \psi=0$.

Since $\theta=0$, we have $x'=x, y'=y$, and the following:

\begin{equation}
\begin{cases}
&f_{00} = f_{02} = f_{20} = f_{22} =\ \exp\left(-\frac{1 + \gamma ^ 2}{2 \sigma ^ 2}\right) \\
&f_{01} = f_{21} =\ \exp\left(-\frac{1}{2 \sigma ^ 2}\right) \\
&f_{10} = f_{12} =\ \exp\left(-\frac{\gamma ^ 2}{2 \sigma ^ 2}\right) \\
&f_{11} = 1\\
\end{cases}
\tag{14.3}
\label{prop2:eqn3}
\end{equation}

We make the following shorthands for conveniency:

\begin{equation}
\begin{split}
h_1 =&\ \exp\left(-\frac{1 + \gamma ^ 2}{2 \sigma ^ 2}\right) \\
h_2 =&\ \exp\left(-\frac{1}{2 \sigma ^ 2}\right) \\
h_3 =&\ \exp\left(-\frac{\gamma ^ 2}{2 \sigma ^ 2}\right) \\
\end{split}
\tag{14.4}
\label{prop2:eqn4}
\end{equation}

From Eqn.\ref{prop2:eqn2}, we can get:

\begin{equation}
(a_{00} + a_{02} + a_{20} + a_{22})h_1 + (a_{01} + a_{21})h_2 + (a_{10} + a_{12})h_3 + a_{11} = 0.
\tag{14.5}
\label{prop2:eqn5}
\end{equation}

The equation above means that, $\exists\ b_0, b_1, b_2, b_3 \in \mathbb{R}$, such that

\begin{equation}
b_0 + b_1 h_1 + b_2 h_2 + b_3 h_3 = 0.
\tag{14.6}
\label{prop2:eqn6}
\end{equation}

Differentiate on both sides of parameter $\gamma$ and get:

\begin{equation}
\begin{split}
&-\frac{\gamma}{\sigma ^ 2} b_1 h_1 - \frac{\gamma}{\sigma ^ 2} b_3 h_3 = 0 \\
\Longrightarrow & b_1 + \exp\left(\frac{\gamma ^ 2}{\sigma ^ 2}\right) b_3 = 0
\end{split}
\tag{14.7}
\label{prop2:eqn7}
\end{equation}

Since Eqn.\ref{prop2:eqn7} holds for $\forall \gamma, \sigma$, which indicates that:

\begin{equation}
b_1 = b_3 = 0
\tag{14.8}
\label{prop2:eqn8}
\end{equation}

In the same way, we can get the following equation from Eqn.\ref{prop2:eqn8}:

\begin{equation}
\begin{split}
&b_2 h_2 + b_0 = 0 \\
&b_0 = b_2 = 0
\end{split}
\tag{14.9}
\label{prop2:eqn9}
\end{equation}

Therefore, we have the following equations:

\begin{equation}
\begin{cases}
&a_{00} + a_{02} + a_{20} + a_{22} =\ 0 \\
&a_{01} + a_{21} =\ 0 \\
&a_{10} + a_{12} =\ 0 \\
&a_{11} =\ 0 \\
\end{cases}
\tag{14.10}
\label{prop2:eqn10}
\end{equation}

\noindent \textbf{\RomanNumeralCaps{2}}\label{prop2:sit2}. $\theta=0,\ \lambda=3,\ \psi=\pi / 3$.

In the same way, we have the following equations:

\begin{equation}
\begin{cases}
&f_{20} = f_{21} = f_{22} =\ 0 \\
&f_{00} = f_{02} =\ \frac{1}{2} h_1 \\
&f_{10} = f_{12} =\ \frac{1}{2} h_3 \\
&f_{01} =\ \frac{1}{2} h_2 \\
&f_{11} =\ \frac{1}{2} \\
\end{cases}
\tag{14.11}
\label{prop2:eqn11}
\end{equation}

And we can get:

\begin{equation}
(a_{00} + a_{02}) h_1 + a_{01} h_2 + (a_{10} + a_{12}) h_3 + a_{11} = 0,
\tag{14.12}
\label{prop2:eqn12}
\end{equation}

which indicates that:

\begin{equation}
\begin{cases}
&a_{00} + a_{02} =\ 0 \\
&a_{01} =\ 0 \\
\end{cases}
\tag{14.13}
\label{prop2:eqn13}
\end{equation}

From Eqn.\ref{prop2:eqn10}, we can get:

\begin{equation}
\begin{cases}
&a_{00} + a_{02} =\ 0 \\
&a_{01} = a_{21} =\ 0 \\
\end{cases}
\tag{14.14}
\label{prop2:eqn14}
\end{equation}

\noindent \textbf{\RomanNumeralCaps{3}}. $\theta=0,\ \lambda=3,\ \psi=-\pi / 3$.

We can get the following equations in the way just the same as discussed in situaltion \RomanNumeralCaps{2}:

\begin{equation}
a_{20} + a_{22} =\ 0
\tag{14.15}
\label{prop2:eqn15}
\end{equation}

\noindent \textbf{\RomanNumeralCaps{4}}. $\theta=\pi / 2,\ \lambda=3,\ \psi=\pm \pi / 3$.

We have $x' = y, y' = x$ this time, and we can get the following equations as the way discussed in situaltion \RomanNumeralCaps{2} \RomanNumeralCaps{3}, :

\begin{equation}
\begin{cases}
&a_{00} + a_{20} =\ 0 \\
&a_{02} + a_{22} =\ 0 \\
&a_{10} = a_{12} = 0 \\
\end{cases}
\tag{14.16}
\label{prop2:eqn16}
\end{equation}

Combine equations \ref{prop2:eqn14} \ref{prop2:eqn15} \ref{prop2:eqn16}, we can get:

\begin{equation}
\begin{cases}
&a_{00} = -a_{02} = a_{22} = -a_{20} = v \\
&a_{01} = a_{10} = a_{12} = a_{21} = 0 \\
\end{cases}
\tag{14.17}
\label{prop2:eqn17}
\end{equation}

\noindent \textbf{\RomanNumeralCaps{5}}. $\theta=\pi / 4,\ \lambda=2,\ \psi=\sqrt{2}\pi$.

We have $x' = \frac{\sqrt{2}}{2}(x + y), y' = \frac{\sqrt{2}}{2}(y - x)$ and the following:

\begin{equation}
\begin{cases}
f_{00} &= \exp\left(-\frac{1}{\sigma ^ 2}\right) \\
f_{02} &= f_{20} =\ \exp\left(-\frac{\gamma ^ 2}{\sigma ^ 2}\right) \\
f_{22} &= \exp\left(-\frac{1}{\sigma ^ 2}\right)\cos(2 \sqrt{2} \pi) \\
\end{cases}
\tag{14.18}
\label{prop2:eqn18}
\end{equation}

Therefore, we have

\begin{equation}
\begin{split}
(1 + \cos(2 \sqrt \pi)) &\exp(-\frac{1}{\sigma ^ 2}) v + 2 \cos(2 \sqrt{2} \pi) \exp(-\frac{\gamma ^ 2}{\sigma ^ 2}) v = 0 \\
\Longrightarrow &\ v = 0 \\
\Longrightarrow &\ a_{00} = a_{02} = a_{20} = a_{22} = 0 \\
\end{split}
\tag{14.19}
\label{prop2:eqn19}
\end{equation}

Combine equations \ref{prop2:eqn10} \ref{prop2:eqn17} \ref{prop2:eqn19}, we can find that $a_{ij} = 0, for i, j = 0, 1, 2$, which means that $I = 0$. Therefore, the assumption that $I \neq 0$ is not true.

For an arbitary sized input $I$, we can focus on the $3 \times 3$ sized submatrix that will do inner product with the convolution kernel and get the same conclusion.

\end{proof}


\begin{proof}[\textbf{Proof of Corollary1:}]
~\\
From Prop.\ref{sec:them:prop2}, the conclusion is obvious.
\end{proof}


\begin{proof}[\textbf{Proof of Theorem1:}]
~\\
Notice that \\
\begin{equation}
\begin{split}
N(\hat{\mathbb{E}}_{\mathcal{S}}[G]-\hat{\mathbb{E}}_{\mathcal{S}}[F]) =& \ \sum_i \tilde{G}_i^2-\tilde{F}_i^2 -2y_i(\tilde{G}_i-\tilde{F}_i) \\
=& \ \sum_i(\tilde{G}_i-\tilde{F}_i)(\tilde{G}_i+\tilde{F}_i-2y_i) \\
\end{split}
\tag{15.1}
\label{proof:theorem1.1}
\end{equation}

Apply absolute value on both sides

\begin{equation}
\begin{split}
N|\hat{\mathbb{E}}_{\mathcal{S}}[G]-\hat{\mathbb{E}}_{\mathcal{S}}[F]| \leq &\ \sum_i |\tilde{G}_i-\tilde{F}_i||\tilde{G}_i+\tilde{F}_i-2y_i| \\
\leq & \ 4\sum_i |\tilde{G}_i-\tilde{F}_i|
\end{split}
\tag{15.2}
\label{proof:theorem1.2}
\end{equation}

The last inequality holds as $|\tilde{F}_i|, |\tilde{G}_i|, |y_i| \le 1$.

We can fix parameters of ordinary CNN, so that there is a mapping between input $I_i$ and output $\tilde{F}_i$, and the mapping function is $\tilde{F}$ as we have defined.

We can also fix parameters of $C_G$, and choose the convolution kernel of $C_G$ that satisfies (\ref{eqn:injective}) since $G$ is a well-defined Geometric Operator CNN, so that $C_G$ is an injective function, which means that $C_G^{-1}$ exists. In the same time, $D_G = FC_{G, 2} \circ \sigma \circ FC_{G, 1}$ can be treated as a one hidden layer neural network.

Define a new hypothesis $\tilde{h} = \sigma \circ F \circ C_G^{-1} \circ \sigma^{-1}$ ranges in $[0, 1]$, according to Lemma.\ref{sec:them:lemma1}, we can find paramters $\{a_{G, k}, b_{G, k}\}, k=1, 2$, such that

\begin{equation}
|\sigma \circ D_G(x) - \tilde{h}(x)| \le \epsilon/4, \quad \forall x.
\tag{15.3}
\label{proof:theorem1.3}
\end{equation}

Replace $x$ by $\sigma \circ C_G(I_i)$ we can get

\begin{equation}
\begin{split}
&|\tilde{G}_i - \tilde{F}_i| \\
=\ &|\tilde{G}(I_i) - \tilde{F}(I_i)| \\
=\ &|\sigma \circ D_G \circ \sigma \circ C_G(I_i) - \sigma \circ F(I_i)| \\
=\ &|\sigma \circ D_G(\sigma \circ C_G(I_i)) - \sigma \circ F \circ C_G^{-1} \circ \sigma^{-1}(\sigma \circ C_G(I_i))| \\
\le\ &\epsilon/4
\end{split}
\tag{15.4}
\label{proof:theorem1.4}
\end{equation}

Combine with (\ref{proof:theorem1.2}), we can get

\begin{equation}
|\hat{\mathbb{E}}_{\mathcal{S}}[G]-\hat{\mathbb{E}}_{\mathcal{S}}[F]| \le\ \frac{4}{N}\sum_i |\tilde{G}_i - \tilde{F}_i| \le\ \frac{4}{N}\frac{N}{4}\epsilon =\ \epsilon
\tag{15.5}
\label{proof:theorem1.5}
\end{equation}

\end{proof}

%

\begin{proof}[\textbf{Proof of Theorem2:}]
~\\
From Theorem.\ref{sec:them:thm1}, we know that $G$ satisfies the following inequality:

\begin{equation}
|\hat{\mathbb{E}}_{\mathcal{S}}[G] - \hat{\mathbb{E}}_{\mathcal{S}}[F]| \leq \epsilon
\tag{16.1}
\label{proof:theorem2.1}
\end{equation}

From Lemma.\ref{sec:them:lemma3}, we know that

\begin{equation}
\begin{split}
\hat{\mathbb{E}}_{\mathcal{D}}[G] \leq &\  \hat{\mathbb{E}}_{\mathcal{S}}[G] + 2\hat{\mathfrak{R}}_{S}^a(\mathcal{G}_f^*)+\sqrt{\frac{log(1/\delta)}{2N}} \\
\leq & \hat{\mathbb{E}}_{\mathcal{S}}[F] + 2\hat{\mathfrak{R}}_{\mathcal{S}}^a(\mathcal{G}_f^*)+\sqrt{\frac{log(1/\delta)}{2N}} + \epsilon
\end{split}
\tag{16.2}
\label{proof:theorem2.2}
\end{equation}

Since $\mathcal{G}_f^* \subset \mathcal{F}$, we have the following inequality from Lemma.\ref{sec:them:lemma2}:

\begin{equation}
\hat{\mathfrak{R}}_{\mathcal{S}}^a(\mathcal{G}_f^*) \leq \hat{\mathfrak{R}}_{\mathcal{S}}^a(\mathcal{F})
\tag{16.3}
\label{proof:theorem2.3}
\end{equation}

Combined with (\ref{proof:theorem2.2}), we have

\begin{equation}
\hat{\mathbb{E}}_{\mathcal{D}}[G] \leq \ \hat{\mathbb{E}}_{\mathcal{S}}[F] + 2\hat{\mathfrak{R}}_{S}^a(\mathcal{F})+\sqrt{\frac{log(1/\delta)}{2N}} + \epsilon
\tag{16.4}
\label{proof:theorem2.4}
\end{equation}

The conclusion is proved!

\end{proof}

%

\begin{proof}[\textbf{Proof of Corollary2:}]
~\\
From Theorem.\ref{sec:them:thm2} and Corollary.\ref{sec:them:coro1}, this conclusion is obvious.
\end{proof}


\begin{proof}[\textbf{Proof of Corollary3:}]
~\\
Let $\mathcal{K}$ be the set of $k_i$ such that the generator function of $k_i$ is $f_{t_i} = f_{t^*}$ and denote the concatenation of all these $k_i$ as $\hat{k}$.

Suppose that there exists an input $x$, satisfies that $x * k_i = 0, i=1, 2, \cdots, od$, then $x * \tilde{k} = 0, \forall \tilde{k} \in \mathcal{K}$. Therefore, $x * \hat{k}$ holds for any paramters. However, it is conflict with Prop.\ref{sec:them:prop2}.

Therefore, the conclusion is proved!
\end{proof}


\bibliographystyle{aaai}
\bibliography{reference}

\end{document}